\documentclass{article}
\usepackage{arxiv}

\usepackage[utf8]{inputenc} % allow utf-8 input
\usepackage[T1]{fontenc}    % use 8-bit T1 fonts
\usepackage{hyperref}       % hyperlinks
\usepackage{url}            % simple URL typesetting
\usepackage{booktabs}       % professional-quality tables
\usepackage{amsfonts}       % blackboard math symbols
\usepackage{nicefrac}       % compact symbols for 1/2, etc.
\usepackage{microtype}      % microtypography
\usepackage{lipsum}
\usepackage{float}
\usepackage{graphicx}
\usepackage{multirow}
\usepackage{array}
\usepackage{comment}
\usepackage{longtable}
\usepackage{ragged2e}
\usepackage{subfig}
\usepackage[table,xcdraw]{xcolor}  % ensure xcolor is loaded with table option
\usepackage{geometry}
\usepackage{tcolorbox} % added to support tcolorbox environment
\usepackage{pifont}
\usepackage{multicol}
 \usepackage{amsmath}
\title{Digital Twins in Additive Manufacturing: A Systematic Review}

\author{
  Md Manjurul Ahsan \\
  Industrial and Systems Engineering\\
  University of Oklahoma\\
  Norman, Oklahoma-73071 \\
  \texttt{ahsan@ou.edu} \\
   \And
  Yingtao Liu \\
  Aerospace and Mechanical Engineering\\
  University of Oklahoma\\
  Norman, Oklahoma-73071\\
  \texttt{yingtao@ou.edu} \\
   \And
  Shivakumar Raman \\
  Industrial and Systems Engineering\\
  University of Oklahoma\\
  Norman, Oklahoma-73071\\
  \texttt{raman@ou.edu} \\
   \And
  Zahed Siddique \\
  Aerospace and Mechanical Engineering\\
  University of Oklahoma\\
  Norman, Oklahoma-73071\\
  \texttt{zsiddique@ou.edu} \\
}

\begin{document}
\maketitle

\begin{abstract}
Digital Twins (DTs) are becoming popular in Additive Manufacturing (AM) due to their ability to create virtual replicas of physical components of AM machines, which helps in real-time production monitoring. Advanced techniques such as Machine Learning (ML), Augmented Reality (AR), and simulation-based models play key roles in developing intelligent and adaptable DTs in manufacturing processes. However, questions remain regarding scalability, the integration of high-quality data, and the computational power required for real-time applications in developing DTs. Understanding the current state of DTs in AM is essential to address these challenges and fully utilize their potential in advancing AM processes. Considering this opportunity, this work aims to provide a comprehensive overview of DTs in AM by addressing the following four research questions: (1) What are the key types of DTs used in AM and their specific applications? (2) What are the recent developments and implementations of DTs? (3) How are DTs employed in process improvement and hybrid manufacturing? (4) How are DTs integrated with Industry 4.0 technologies? By discussing current applications and techniques, we aim to offer a better understanding and potential future research directions for researchers and practitioners in AM and DTs.
\end{abstract}

\section*{Abbreviations}
\begin{multicols}{2}
\begin{itemize}
    \item \textbf{AD:} Axiomatic Design
    \item \textbf{AI:} Artificial Intelligence
    \item \textbf{AM:} Additive Manufacturing
    \item \textbf{AR:} Augmented Reality
    \item \textbf{CNC:} Computer Numerical Control
    \item \textbf{CAD:} Computer-Aided Design
    \item \textbf{DED:} Direct Energy Deposition
    \item \textbf{DfA:} Design for Assembly
    \item \textbf{DfAM:} Design for Additive Manufacturing
    \item \textbf{DLP:} Digital Light Processing
    \item \textbf{DPF:} Digital Part File
    \item \textbf{DT:} Digital Twin
    \item \textbf{DTs:} Digital Twins
    \item \textbf{EBM:} Electron Beam Melting
    \item \textbf{FDM:} Fused Deposition Modeling
    \item \textbf{FEM:} Finite Element Method
    \item \textbf{FPS:} Frames Per Second
    \item \textbf{GUI:} Graphical User Interface
    \item \textbf{HCPS:} Human-Cyber-Physical Systems
    \item \textbf{I5.0:} Industry 5.0
    \item \textbf{IoT:} Internet of Things
    \item \textbf{KPIs:} Key Performance Indicators
    \item \textbf{L-DED:} Laser-Directed Energy Deposition
    \item \textbf{LPBF:} Laser Powder Bed Fusion
    \item \textbf{ML:} Machine Learning
    \item \textbf{MFDT:} Multisensor Fusion-based Digital Twin
    \item \textbf{PBF:} Powder Bed Fusion
    \item \textbf{PINNs:} Physics-Informed Neural Networks
    \item \textbf{RNN:} Recurrent Neural Network
    \item \textbf{RMSE:} Root Mean Square Error
    \item \textbf{SEM:} Scanning Electron Microscopy
    \item \textbf{SLA:} Stereolithography
    \item \textbf{SLS:} Selective Laser Sintering
    \item \textbf{SLM:} Selective Laser Melting
    \item \textbf{TDABC:} Time-Driven Activity-Based Costing
    \item \textbf{VACCY:} Volume Approximation by Cumulated Cylinders
    \item \textbf{VQVAE-GAN:} Vector Quantized Variational AutoEncoder and Generative Adversarial Network
    \item \textbf{VR:} Virtual Reality
    \item \textbf{WAAM:} Wire Arc Additive Manufacturing
\end{itemize}
\end{multicols}

\section{Introduction}\label{sec1}
 \textbf{What is Additive Manufacturing?}
 
\begin{figure}[h]
    \centering
    \includegraphics[width=\textwidth]{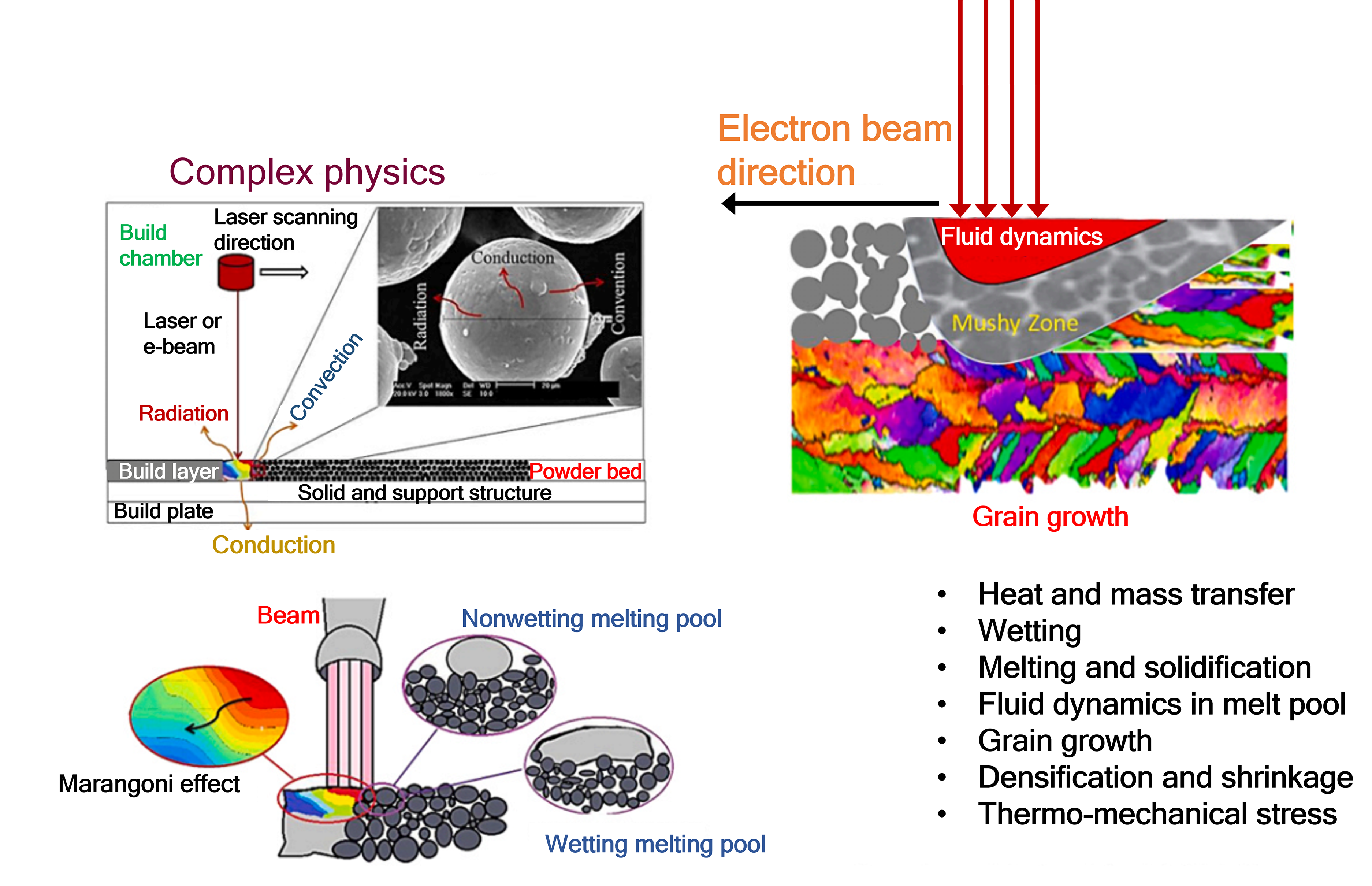}
    \caption{\textbf{The multiscale nature of the powder-bed metal AM process, showing various properties and interactions across different length and time scales~\cite{ladani2021additive}.}}
    \label{fig:addifirst}
\end{figure}

\textbf{Additive Manufacturing} (AM), also known as 3D printing, produces three-dimensional objects from a digital file by layering material incrementally. Unlike traditional methods, which usually remove material from a larger block, AM constructs objects directly from a digital design~\cite{gibson2014additive, li2018additive, weller2015economic}. This approach enables the creation of intricate geometries that are challenging or impossible to achieve with conventional techniques. AM can work with a variety of materials, such as plastics, metals, and ceramics~\cite{petrovic2011additive, guo2013additive}. Typically, the digital design is created using Computer-Aided Design (CAD) software, which is then sliced into thin layers for the AM machine to read and construct sequentially~\cite{kruth1998progress}. \textbf{Figure~\ref{fig:addifirst}} illustrates an example of the AM process in a powder bed metal scenario.
\begin{tcolorbox}[colback=yellow!10!white, colframe=yellow!80!black, title=Benefits of AM]
The benefits of AM include:
\begin{itemize}
    \item \textbf{Customization:} The ability to create customized products tailored to specific needs~\cite{gibson2014additive, weller2015economic}.
    \item \textbf{Material Efficiency:} Reducing waste by using only the material necessary for the build~\cite{petrovic2011additive}.
    \item \textbf{Complex Geometries:} Creating complex and intricate designs that are not possible with traditional methods~\cite{guo2013additive}.
    \item \textbf{Rapid Prototyping:} Quickly producing prototypes for testing and validation, speeding up the product development cycle~\cite{kruth1998progress}.
\end{itemize}
\end{tcolorbox}

\textbf{What are the types of AM?}

\begin{figure}[htbp]
    \centering
    \includegraphics[width=.9\textwidth]{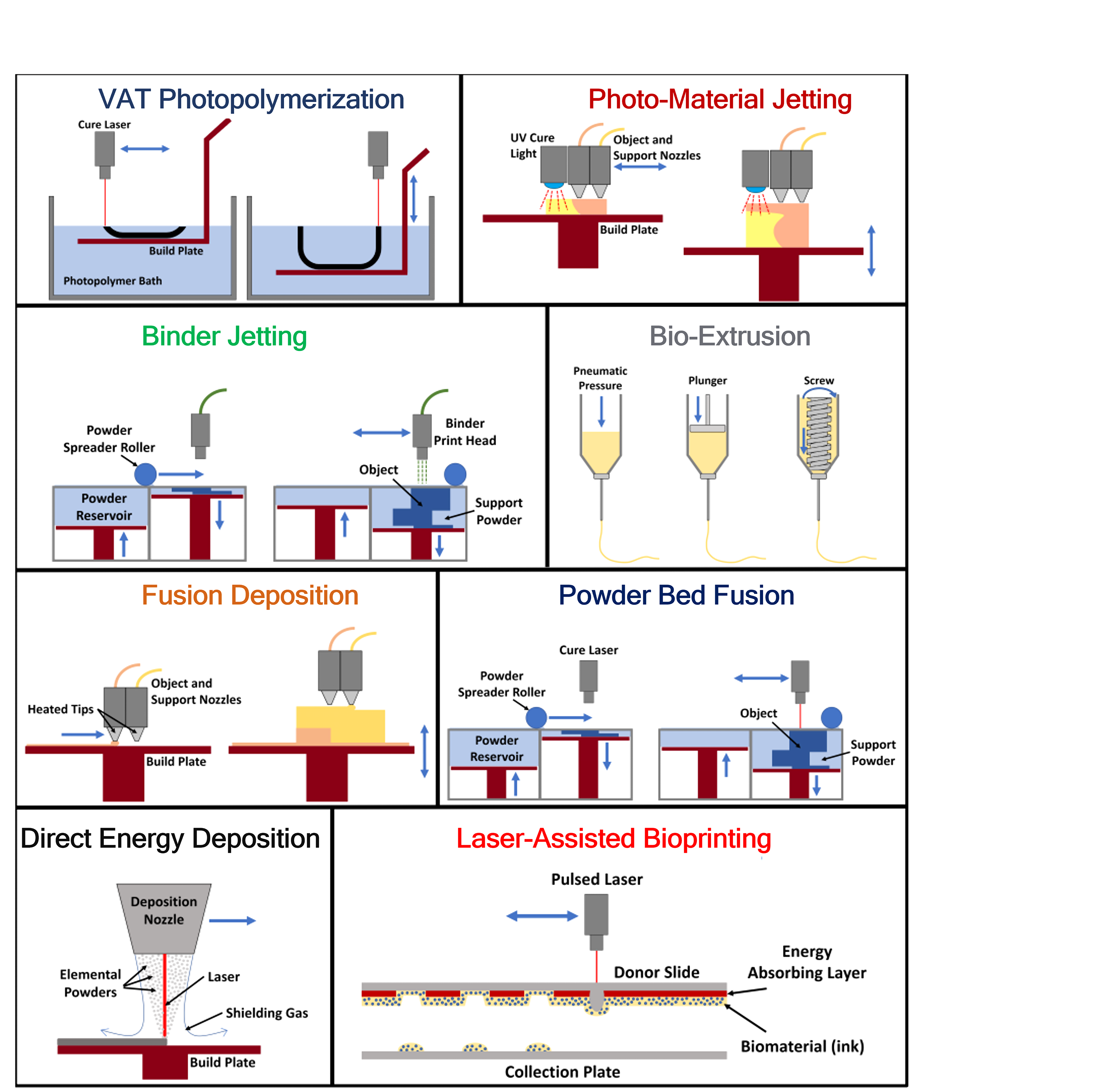}
    \caption{\textbf{Different types of AM techniques~\cite{francis2018additive}.}}
    \label{fig:ammetaltype}
\end{figure}

There are several types of AM technologies, each with its unique process and applications (as shown in \textbf{Figure~\ref{fig:ammetaltype})}. The most common types include:
\begin{itemize}
\item[\ding{113}] \textbf{Stereolithography (SLA):} SLA employs a laser to cure liquid resin into hardened plastic through a layer-by-layer process. This method is renowned for creating parts with high resolution and smooth surface finishes~\cite{hull2013stereolithography}.
    \item[\ding{113}] \textbf{Fused Deposition Modeling (FDM):} FDM extrudes thermoplastic filament through a heated nozzle, which is deposited layer by layer to create a part. It is widely used due to its simplicity and cost-effectiveness~\cite{crump1992app}.
    
    \item[\ding{113}] \textbf{Selective Laser Sintering (SLS):} SLS uses a laser to sinter powdered material, typically nylon or other polymers, into a solid structure. This method is known for its ability to create strong and durable parts without the need for support structures~\cite{deckard1989method}.
    \item[\ding{113}] \textbf{Digital Light Processing (DLP):} This method is similar to SLA but uses a digital light projector to cure photopolymer resin. It can quickly produce high-resolution parts~\cite{bártolo2011stereolithography}.
    \item[\ding{113}] \textbf{Binder Jetting:} In this process, a liquid binding agent is deposited over a powder bed to bond the material and form a solid part. It's suitable for metals, ceramics, and even sand~\cite{mostafaei2016binder}.
    \item[\ding{113}] \textbf{Material Jetting:} This technique involves depositing droplets of build material layer by layer, which are then cured by UV light. It allows for high precision and can print multiple materials at once~\cite{derby2010inkjet}.
    \item[\ding{113}] \textbf{Powder Bed Fusion (PBF):} PBF includes technologies like Selective Laser Melting (SLM) and Electron Beam Melting (EBM). These use a laser or electron beam to melt and fuse powder particles together~\cite{herzog2016additive}.
    \item[\ding{113}] \textbf{Direct Energy Deposition (DED):} DED uses thermal energy, such as a laser or electron beam, to fuse materials by melting them as they are deposited. This method is used for repairing and adding material to existing components~\cite{debroy2018additive}.
    \item[\ding{113}] \textbf{Sheet Lamination:} This method involves stacking and bonding sheets of material, which are then cut to shape using a laser or another cutting tool. It is known for its speed and ability to use a wide range of materials~\cite{kurdi2014sheet}.
\end{itemize}

\textbf{What are the processes of AM?}

\begin{figure}
    \centering
    \includegraphics[width=\textwidth]{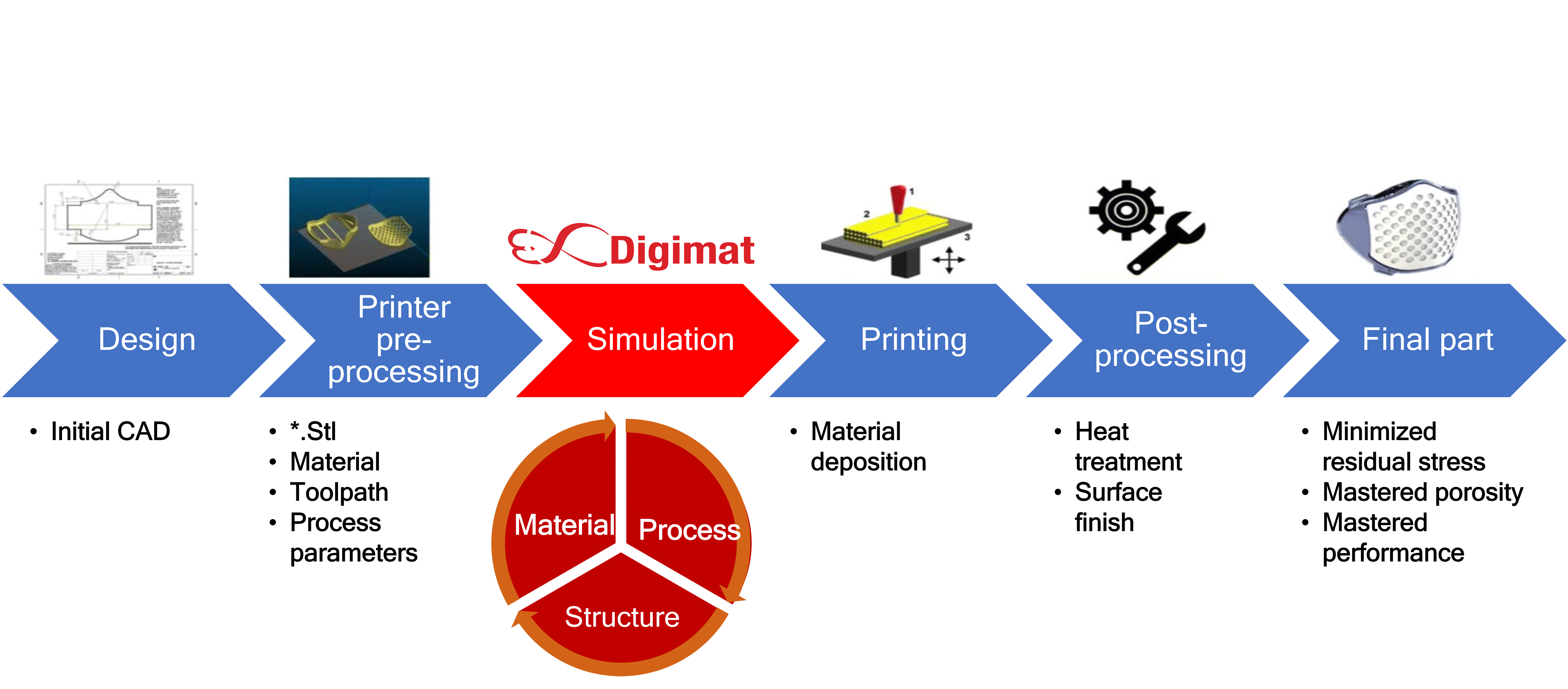}
   \caption{\textbf{Different stages of the AM process in bio-material domains~\cite{tarfaoui2020additive}.}}
    \label{fig:am1}
\end{figure}

AM involves a series of operations that build objects layer by layer from digital models. These processes can be broadly classified into the following categories:

\begin{enumerate}
    \item \textbf{Design.} The process begins with creating a digital 3D model of the object using CAD software. This model serves as the blueprint for the AM process.
    \item \textbf{Slicing.} The digital model is then sliced into thin horizontal layers using slicing software. This step translates the 3D model into instructions that the AM machine can follow to build the object layer by layer~\cite{gibson2014additive}.
    \item \textbf{Material Preparation.} The appropriate material for the chosen AM process is prepared. Materials can range from thermoplastics and photopolymers to metals and ceramics, depending on the specific AM technology being used~\cite{kruth1998progress}.
    \item \textbf{Printing.} The AM machine follows the instructions from the sliced model to add material layer by layer. This step varies significantly between different AM technologies as mentioned earlier.
    \item \textbf{Post-Processing.} Once the object is printed, it may require post-processing to achieve the desired finish and mechanical properties. This can include removing support structures, sanding, polishing, heat treatment, or other finishing processes~\cite{gibson2014additive}.
    \item \textbf{Inspection and Testing.} The final step involves inspecting and testing the manufactured object to ensure it meets the required specifications and quality standards. This can involve visual inspections, dimensional measurements, mechanical testing, and other quality control procedures~\cite{coleman2020sensitivity}. \textbf{Figure~\ref{fig:am1}} illustrates the steps of the AM process in biomaterials.
    
\end{enumerate}

\textbf{What is Digital Twin?}

\begin{figure}[h]
    \centering
    \includegraphics{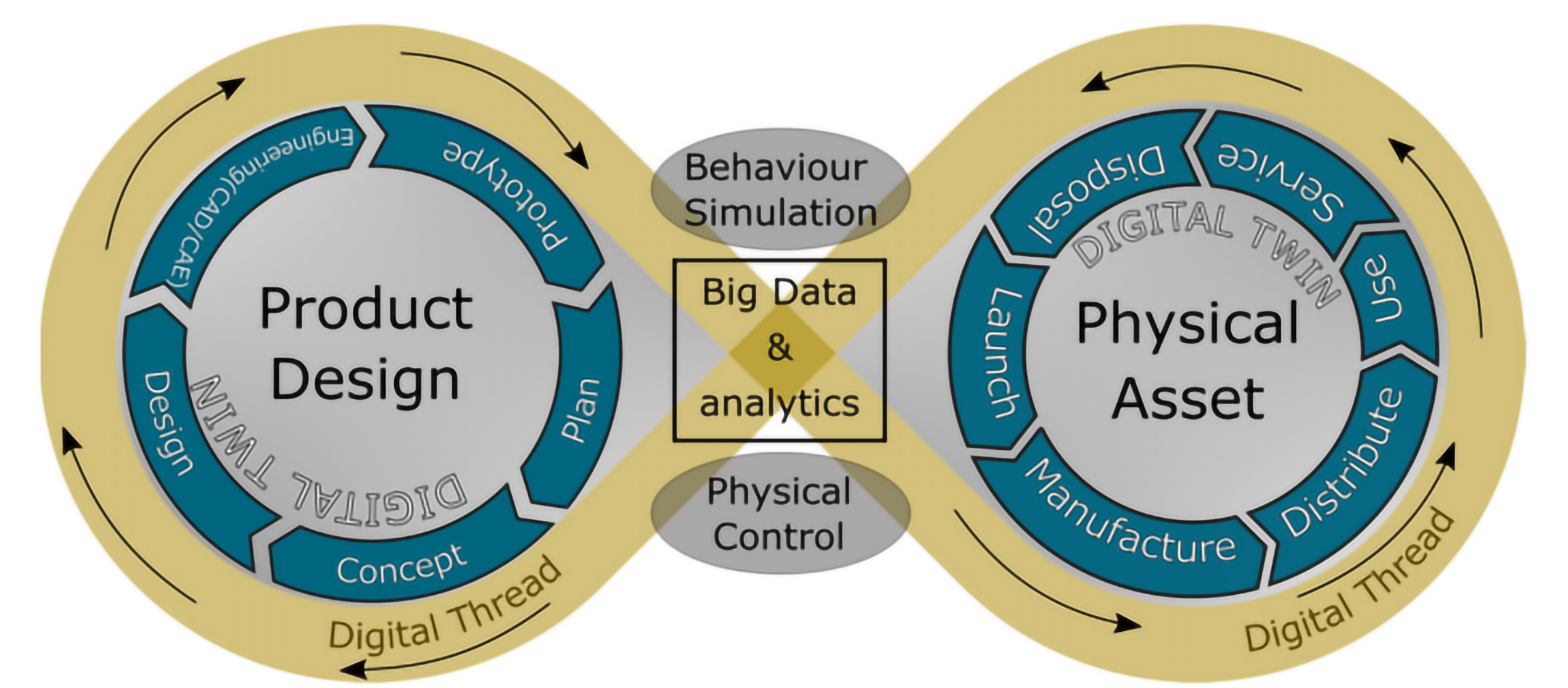}
\caption{\textbf{An efficient product data management framework employs a multi-layered structure within a DT. This approach uses an iterative loop, allowing developers to continuously refine and enhance the model throughout different development stages~\cite{pang2021developing}.}}
    \label{fig:dtframe}
\end{figure}

A Digital Twin (DT) is a virtual model of a physical entity or system that simulates, analyzes, and optimizes its real-world counterpart (Figure~\ref{fig:dtframe}). It involves creating an accurate digital model that mirrors the characteristics, conditions, and behaviors of the physical object, allowing for ongoing monitoring and predictive analysis. Digital Twins (DTs) use data from sensors and Internet of Things (IoT) devices to update the virtual model in real time, providing insights to improve performance, reduce downtime, and optimize operations~\cite{grieves2014digital}. 

DTs are used in various fields like manufacturing, healthcare, and smart cities, where they improve decision-making and operational efficiency~\cite{tao2018digital}. By integrating Machine Learning (ML) and data analytics, DTs offer a complete approach to managing complex systems and processes, driving innovation and efficiency~\cite{fuller2020digital}.

DTs in AM offer a new way to increase efficiency. DTs enable real-time monitoring, simulation, and optimization by creating a digital copy of the AM process, which ensures each phase meets specific standards. This helps manufacturers predict and solve potential issues before they arise, reduce waste, and minimize downtime~\cite{tao2018digital}.

One of the major benefits of DTs in AM is virtual testing and validation. Within this kind of digital environment, engineers are able to perform experiments regarding different materials, designs, or process parameters. This accelerates research and development. DTs provide simulations of the overall build process to find out the best conditions and early defect detection for complex and precise component creation~\cite{fuller2020digital}.

DTs also provide predictive maintenance for AM equipment. By continuously gathering data for analysis from sensors within the machines, DTs are capable of predicting equipment failure and scheduling maintenance in advance in order to prevent problems from emerging. This extends the life of machinery and increases productivity by having the machines running efficiently with reduced occurrences of breakdown~\cite{qi2019enabling}.

Moreover, DTs in AM support product customization and scaling. They help optimize process parameters to customer demand and hence can produce customized products at high volumes. This flexibility of production is specially important for producing custom-made components as required in industries like aerospace and healthcare~\cite{zhang2020smart}.

\subsection{Motivation and uniqueness of this survey}
This review paper on DTs in AM is driven by the rapid advancements and the increasing complexity of integrating DTs into these processes. Our aim is to address current challenges and limitations, providing a clear understanding of how DTs can improve efficiency, precision, and cost-effectiveness in AM. We seek to consolidate existing knowledge, highlight significant research gaps, and propose future research directions. By offering a thorough analysis, we hope to promote wider adoption of DTs technology in the industry, thereby improving manufacturing processes and guiding researchers and industry stakeholders towards more effective implementations. In this review, we address four major questions:
\begin{enumerate}
\item What are the key types of DTs used in AM and their specific applications? (Section~\ref{dt4})
    \item What are the recent developments and implementations of DTs? (Section~\ref{dt1})
    \item How are DTs being employed in process improvement and hybrid manufacturing? (Section~\ref{dt2})
    \item How are DTs integrated with Industry 4.0 technologies? (Section~\ref{dt3})
    
\end{enumerate}

\begin{figure}
%\vspace{-5mm}
    \centering
    \includegraphics[width=.9\textwidth]{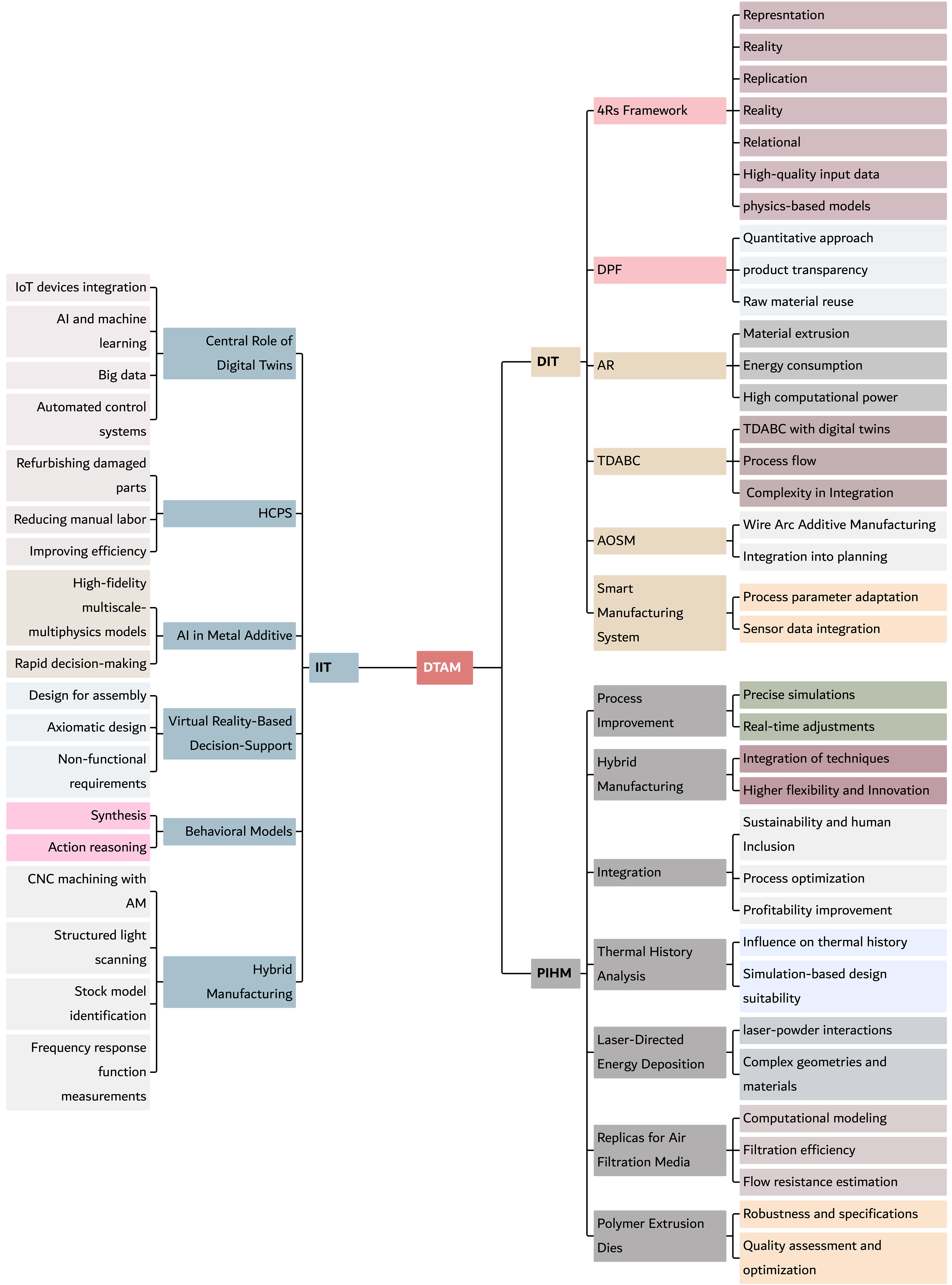}
   % \vspace{-30mm}
    \caption{\textbf{DTs in AM: a framework based on the reference literature; Digital Twins in AM (DTAM), Digital Twins in Industry Technologies (DIT), Process Improvement and Hybrid Manufacturing (PIHM), and Integration of Industry Technologies (IIT), Digital Part File (DPF), Augmented Reality (AR), Time-Driven Activity-Based Costing (TDABC), Adaptive Online Simulation Models (AOSM).}}
    \label{fig:dttttt}
\end{figure}

Our overall contributions are summarized as follows:
\begin{itemize}
    \item[\ding{113}] This review provides a comprehensive overview of the use of DTs in various aspects of AM, including product, process, system, and service DTs.
    \item[\ding{113}] The review highlights significant innovations and methodologies that improve DT performance in AM, focusing on real-time monitoring, optimization, defect prediction, and cost analysis.
    \item[\ding{113}] Key challenges such as data quality, computational demands, scalability issues, and technological integration complexities are addressed, with suggestions for overcoming these limitations.
    \item[\ding{113}] Potential research opportunities are outlined, emphasizing advanced data integration, computational methods, sustainable practices, Augmented Reality (AR) and Virtual Reality (VR) applications, and integration with Industry 4.0 principles. \textbf{Figure~\ref{fig:dttttt}} illustrates the framework of DTs in AM, based on the referenced literature used during this study, discussed in Sections~\ref{dt1}-~\ref{dt4}.
 
\end{itemize}

\subsection{Search strategy}
 Data were sourced from Scopus, initially identifying 337 articles using the title, abstract, and keywords with the search terms "digital twin*" AND "additive manufacturing". 

\begin{figure}[h]
    \centering
    \includegraphics[width=\textwidth]{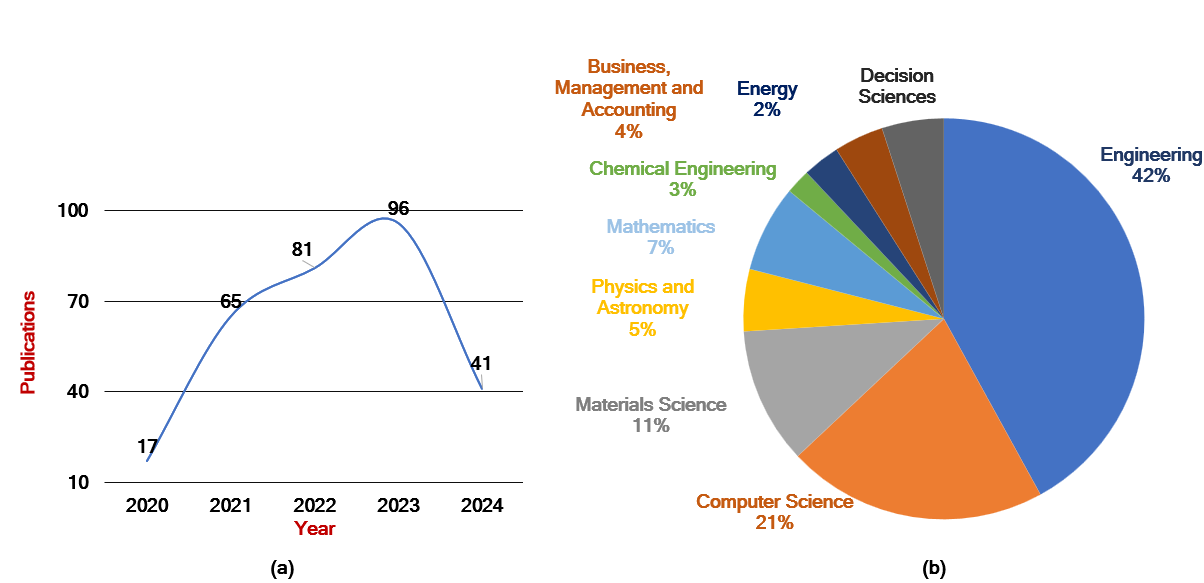}
    \caption{\textbf{This survey examines the most recent advancements in DT, particularly emphasizing the AM sector. It presents statistics on (a) the number of papers published in the past five years on DTs and (b) the distribution percentage of these papers across various domains. The plots illustrate a steady increase in recent literature.}}
    \label{fig:pubyear}
\end{figure}

\textbf{Figure~\ref{fig:pubyear}} shows the number of papers published on DTs in AM over the past five years. \textbf{Figure~\ref{fig:pubyear}(a)} depicts a steady rise in research from 2020 to 2023, with publications increasing from 33 to 85. In 2024, 43 papers have been published so far, showing ongoing research activity. \textbf{Figure~\ref{fig:pubyear}(b)} categorizes these publications by subject. Engineering leads with 42\%, while Energy has the least at 2\%. These statistics emphasize the focus on technical and engineering aspects of DTs in AM, with growing interest in other fields.

Limiting the search to English-language, peer-reviewed, and open-access papers from 2020 to 2024 reduced the count to 120. One researcher (Z.S.) imported these 120 articles into Excel for detailed analysis. Duplicates were removed using Excel's tools. Two independent reviewers (M.A. and Z.S.) evaluated the titles and abstracts, identifying 35 relevant papers. Additionally, 10 more important papers were included, resulting in a total of 45 papers across various fields. We recognize that there may be other important research in the field that we did not cover in our review. Nonetheless, our objective was to provide a comprehensive overview of the most significant and impactful studies.

\section{DTs in AM}

\begin{figure}
    \centering
   \includegraphics[width=\textwidth]{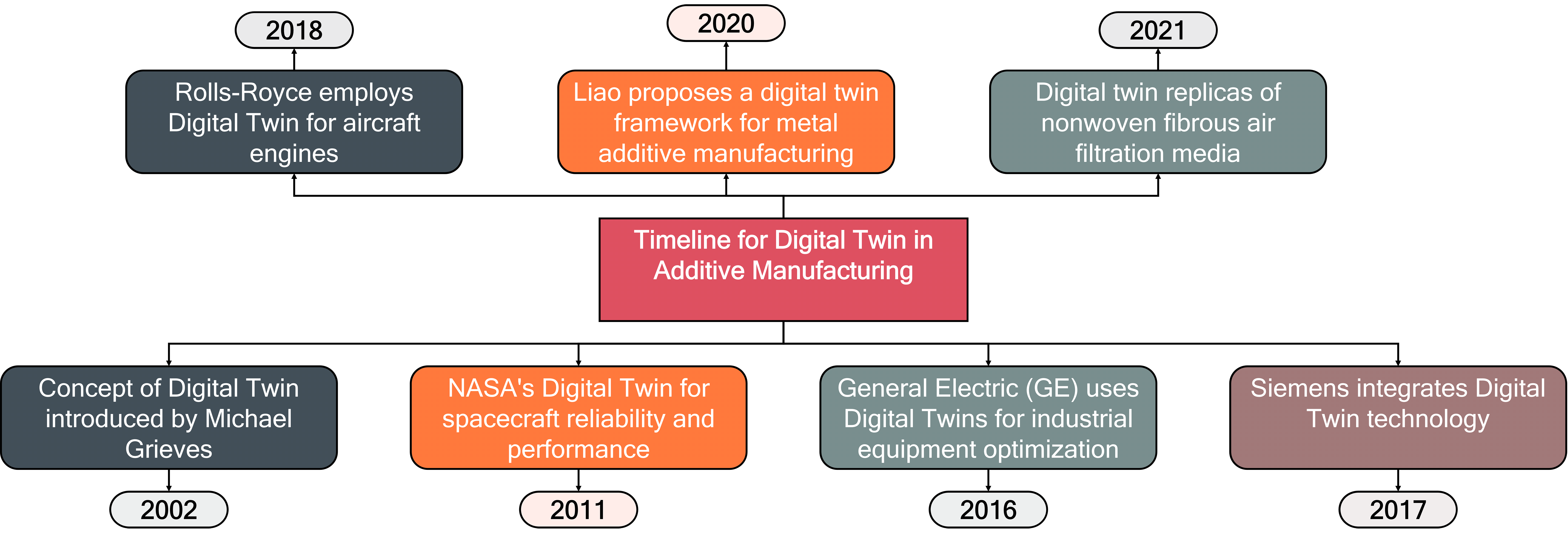}
    \caption{\textbf{Time frame of DTs over the years, from conceptualization to significant integrations and advancements.}}
    \label{fig:dtti}
\end{figure}

In AM, DTs play a crucial role in developing various aspects of the manufacturing process, from design to production and maintenance. \textbf{Figure~\ref{fig:dtti}} illustrates the different stages of DTs and their evolution over the years. From the figure, it can be observed that the concept of DTs was initially introduced by Michael Grieves~\cite{grieves2014digital} and later adopted by several agencies such as NASA~\cite{glaessgen2012digital}, General Electric~\cite{tao2018digital}, Siemens, and others.

The major types of DTs used in AM are \textbf{Product DTs}, \textbf{Process DTs}, and \textbf{System DTs}. Each type plays a unique role, improving the efficiency and effectiveness of the manufacturing process.

\begin{itemize}
    \item[\ding{113}]  \textbf{Product DTs.} In product DTs, designers can test different iterations to arrive at a reshaped final product and forecast its performance in various conditions. Therefore, it helps in reducing design flaws at an early stage, involving less use of physical prototypes, and increasing development speed~\cite{grieves2016origins}.

    \item[\ding{113}]  \textbf{Process DTs.}  Process DTs are mainly useful for models where production workflow layering, temperature distribution, and material flow need to be monitored. These twins provide an opportunity for real-time traceability and control of the manufacturing process, enabling necessary adjustments toward the optimization of production~\cite{tao2018digital}.
   \item[\ding{113}]  \textbf{System DTs.} System DTs integrate multiple product and process twins to provide a holistic view of the manufacturing system. In AM, these twins can monitor and optimize the entire production line, from raw material handling to the final product assembly. System DTs are particularly useful for identifying bottlenecks, improving workflow efficiency, and ensuring that all components work seamlessly together \cite{fuller2020digital}.
   \item[\ding{113}] \textbf{Service DTs.} Service DTs are used for the operational phase of the manufactured components. They provide real-time monitoring and predictive maintenance by collecting data from sensors embedded in the components. This data helps in predicting potential failures and scheduling maintenance activities proactively, thus extending the lifecycle of the components and reducing downtime \cite{boschert2016digital}.
    \item[\ding{113}] \textbf{Hybrid DTs.} Hybrid DTs combine elements of product, process, and system twins to offer a comprehensive solution for managing the entire manufacturing lifecycle. In AM, hybrid twins are used to simultaneously optimize product design, manufacturing processes, and system operations. They offer a unified platform to manage complex manufacturing workflows~\cite{tao2019make}. \textbf{Table~\ref{tab:dttypes}} summarizes the different types of DTs, their applications, advantages, and limitations as well.
\end{itemize}

\begin{table}[h!]
\small
\centering
\caption{\textbf{Different types of DTs in AM.}}
\label{tab:dttypes}
\label{tab:types_of_digital_twins}
\begin{tabular}{p{2cm}p{5cm}p{4cm}p{4cm}}
\toprule
\rowcolor{gray!20} \multicolumn{1}{c}{\textbf{Types of DTs}} & \multicolumn{1}{c}{\textbf{Applications}} & \multicolumn{1}{c}{\textbf{Pros}} & \multicolumn{1}{c}{\textbf{Cons}} \\ \midrule
\rowcolor{green!20} \multirow{5}{*}{\textbf{Product DTs}} & 
\begin{itemize} 
    \item Design optimization 
    \item Lifecycle management 
    \item Performance prediction 
\end{itemize} & 
\begin{itemize} 
    \item Improved product quality 
    \item Reduced time-to-market 
\end{itemize} & 
\begin{itemize} 
    \item High initial setup cost 
    \item Complexity in integration 
\end{itemize} \\ 
\hline
\rowcolor{blue!20} \multirow{5}{*}{\textbf{Process DTs}} & 
\begin{itemize} 
    \item Process optimization 
    \item Real-time monitoring 
    \item Predictive maintenance 
\end{itemize} & 
\begin{itemize} 
    \item Enhanced process efficiency 
    \item Reduced downtime 
\end{itemize} & 
\begin{itemize} 
    \item Requires high-quality data 
    \item Potential cybersecurity risks 
\end{itemize} \\ 
\hline
\rowcolor{orange!20} \multirow{5}{*}{\textbf{System DTs}} & 
\begin{itemize} 
    \item System-level performance analysis 
    \item Resource allocation 
    \item System integration 
\end{itemize} & 
\begin{itemize} 
    \item Holistic view of the system 
    \item Better decision-making 
\end{itemize} & 
\begin{itemize} 
    \item Complexity in modeling 
    \item High computational requirements 
\end{itemize} \\ 
\hline
\rowcolor{pink!20} \multirow{5}{*}{\textbf{Service DTs}} & 
\begin{itemize} 
    \item Real-time monitoring 
    \item Predictive maintenance 
    \item Lifecycle management 
\end{itemize} & 
\begin{itemize} 
    \item Extended component lifespan 
    \item Reduced downtime 
\end{itemize} & 
\begin{itemize} 
    \item Requires high-quality sensor data 
    \item Integration complexity 
\end{itemize} \\ 
\hline
\rowcolor{yellow!20} \multirow{5}{*}{\textbf{Hybrid DTs}} & 
\begin{itemize} 
    \item Combining product and process insights 
    \item End-to-end optimization 
\end{itemize} & 
\begin{itemize} 
    \item Comprehensive insights 
    \item Maximized efficiency 
\end{itemize} & 
\begin{itemize} 
    \item Very complex to implement 
    \item Requires advanced expertise 
\end{itemize} \\ 
\hline
\end{tabular}
\end{table}

\subsection{Steps to Implement DTs in AM}

While there is no strict policy for implementing DTs in AM, there are some general steps required to develop DT models. Here, we considered DTs in Metal AM incorporating ML and physics-based neural networks~\cite{tao2018digital, grieves2014digital}. The general steps of DTs include:

\ding{227} \textbf{Data Collection}

\textit{Sensor Integration.} Equip the AM machine with sensors to collect real-time data on various parameters such as temperature, pressure, and laser power. This is crucial for creating an accurate and detailed DTs. The data collected can be represented as:
\begin{equation}
\mathbf{X} = \{x_1, x_2, \ldots, x_n\}
\end{equation}
where $\mathbf{X}$ represents the set of all collected sensor data.

\textit{Environment and Material Monitoring.} Monitor on environmental factors such as humidity and temperature, along with material properties like powder particle size and flow rate. This ensures the manufacturing process operates under ideal conditions, allowing for quick detection and correction of any deviations.

\ding{227} \textbf{Data Preprocessing}

\textit{Noise Reduction.} Apply filtering techniques to eliminate noise from the collected data. This guarantees the data is clean and precise for further analysis.
\begin{equation}
\mathbf{X}_{\text{filtered}} = G(\mathbf{X})
\end{equation}
where $\mathbf{X}_{\text{filtered}}$ denotes the sensor data after filtering and $G(\mathbf{X})$ represents the Gaussian filter applied to the data.

\textit{Normalization.} Standardize the data to ensure it remains consistent across various measurements and conditions.
\begin{equation}
x' = \frac{x - \min(\mathbf{X})}{\max(\mathbf{X}) - \min(\mathbf{X})}
\end{equation}
where $x'$ represents the normalized data.

\ding{227} \textbf{Model Development}

\textit{Physics-Based Modeling.} Create physics-based models to simulate the AM process. These models use principles like heat transfer and fluid dynamics to predict outcomes.
\begin{equation}
\frac{\partial T}{\partial t} = \alpha \nabla^2 T
\end{equation}
where $T$ represents the temperature, $t$ represents time, and $\alpha$ is the thermal diffusivity.

\textit{Machine Learning (ML) Algorithms.} Train ML models, such as Neural Networks (NN), on historical data to predict process outcomes and optimize parameters.
\begin{equation}
\hat{y} = f(\mathbf{X}; \theta)
\end{equation}
where $\hat{y}$ represents the predicted output, $\mathbf{X}$ represents the input data, $f(\mathbf{X}; \theta)$ represents the NN model, and $\theta$ represents the model parameters.

\textit{Physics-Informed Neural Networks (PINNs).} Integrate physics-based constraints into NN to improve prediction accuracy.
\begin{equation}
\mathcal{L} = \| f(\mathbf{X}; \theta) - y \|^2 + \lambda \left\| \frac{\partial T}{\partial t} - \alpha \nabla^2 T \right\|^2
\end{equation}
where $\mathcal{L}$ represents the loss function, $y$ represents the actual output, and $\lambda$ is a regularization parameter.

\ding{227} \textbf{DT Creation}

\textit{Virtual Model Construction.} Develop a virtual model of the AM process using the established models. This digital model serves as the DT, mirroring the physical process in a digital environment.

\textit{Real-Time Synchronization.} Continuously synchronize the virtual model with real-time data from the physical process to ensure that the DTs remain an accurate representation of the actual manufacturing process.
\begin{equation}
\mathbf{X}_{\text{virtual}} \leftarrow \mathbf{X}_{\text{real-time}}
\end{equation}
where $\mathbf{X}_{\text{virtual}}$ represents the data of the virtual model and $\mathbf{X}_{\text{real-time}}$ represents the real-time data from the physical process.

\ding{227} \textbf{Simulation and Optimization}

\textit{Process Simulation.} Simulate the AM process using the DTs to predict outcomes and identify potential issues.

\textit{Parameter Optimization.} Use optimization algorithms to identify the best process parameters for achieving the desired outcomes.
\begin{equation}
\theta_{t+1} = \theta_t - \eta \nabla \mathcal{L}
\end{equation}
where $\theta_{t+1}$ represents the updated parameters, $\theta_t$ represents the current parameters, $\eta$ is the learning rate, and $\nabla \mathcal{L}$ is the gradient of the loss function.

\ding{227} \textbf{Real-Time Monitoring and Control}

\textit{Anomaly Detection.} Use the DTs to detect anomalies in real-time and trigger alerts when deviations from the expected process occur.
\begin{equation}
\text{if } |x_t - \mu| > 3\sigma, \text{ flag\ an\ anomaly}
\end{equation}
where $x_t$ represents the current data point, $\mu$ is the mean, and $\sigma$ is the standard deviation.

\textit{Feedback Loop.} Implement a feedback loop where the DTs inform the physical process to adjust parameters on-the-fly, ensuring that the manufacturing process remains within optimal conditions.
\begin{equation}
P_{\text{laser}} \leftarrow P_{\text{optimal}}
\end{equation}
where $P_{\text{laser}}$ represents the laser power and $P_{\text{optimal}}$ represents the optimized laser power.

\ding{227} \textbf{Post-Process Analysis}

\textit{Quality Assurance.} Use the DTs to analyze the final product quality, ensuring that it meets the required specifications and standards.

\begin{figure}
    \centering
    \includegraphics[width=\textwidth]{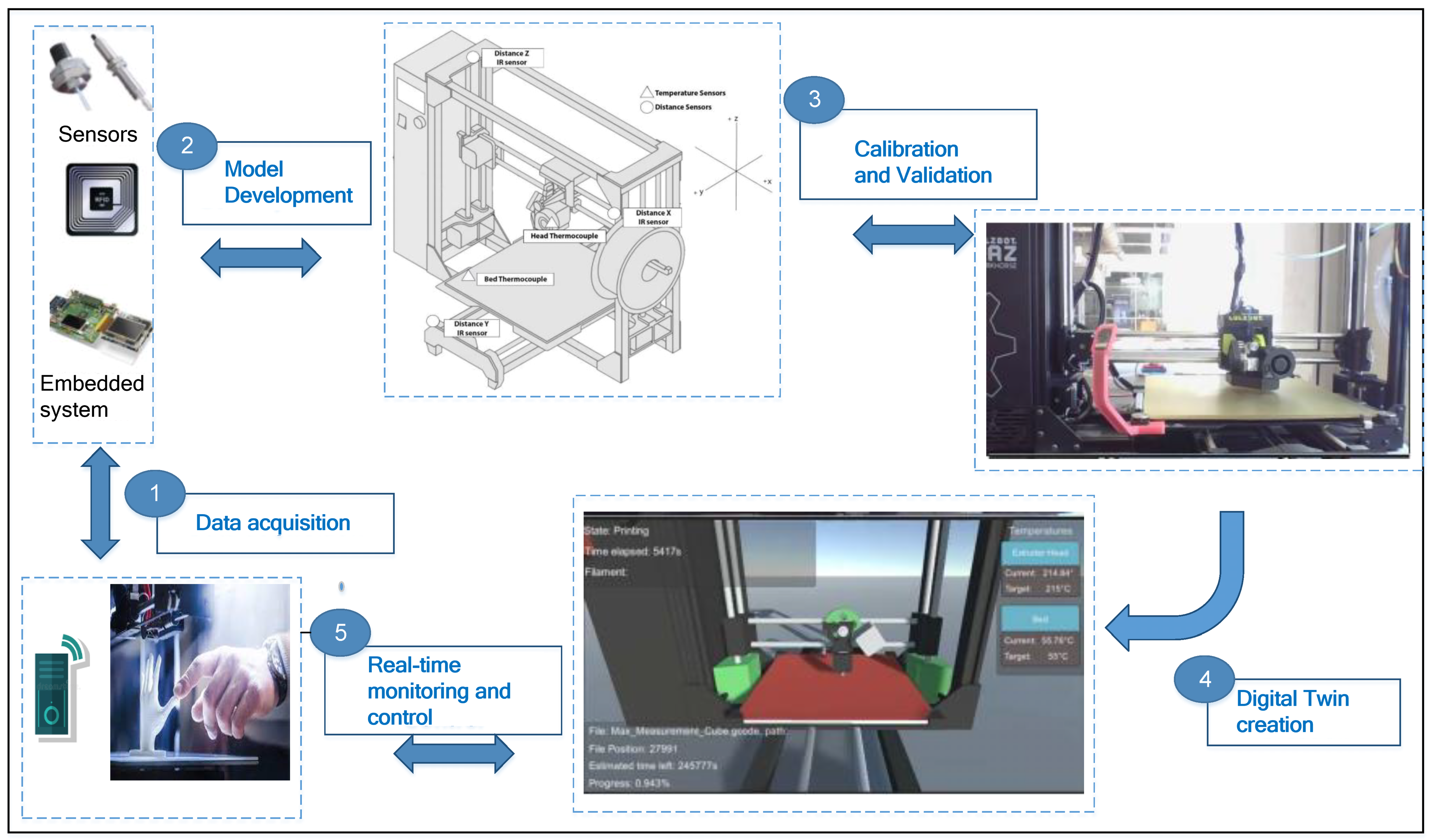}
    \caption{\textbf{Different stages of a Digital Twin system in the additive manufacturing process~\cite{pantelidakis2022digital,gunasegaram2021case,ben2024digital}.}}
    \label{fig:dtsam}
\end{figure}

\textit{Continuous Improvement.} Continuously update the DTs with new data to improve model accuracy and process efficiency.
\begin{equation}
\theta \leftarrow \arg\min_\theta \mathcal{L}(\mathbf{X}_{\text{new}}, y_{\text{new}})
\end{equation}
where $\theta$ represents the model parameters, $\mathcal{L}$ is the loss function, $\mathbf{X}_{\text{new}}$ is the new input data, and $y_{\text{new}}$ is the new output data.
Figure~\ref{fig:dtsam} illustrates some of the major steps in developing digital twins in the additive manufacturing process.

\subsection{DTs with Artificial Intelligence (AI)}
DTs in AM increasingly incorporate AI and ML techniques to enhance control, optimize performance, and improve product quality. AI/ML algorithms process big data from sensors in AM machines. For instance, regression and clustering techniques forecast powder behavior and classify types based on their properties, ensuring precise material simulations \cite{tao2018digital}. Optimization algorithms identify the best machine settings for quality control, while Physics-based Neural Networks predict accurate thermal profiles, capturing critical thermal properties affecting materials \cite{grieves2014digital}. Reinforcement Learning predict the mechanical properties of printed parts, incorporating physics-based constraints for realistic behavior. Real-time data analysis and anomaly detection offer continuous feedback during printing, enabling immediate corrections and maintaining consistent quality. Furthermore, Convolutional Neural Networks inspect each manufacturing layer for defects, ensuring layer-by-layer quality control. Predictive analytics schedules maintenance before failures, prolongs equipment lifespan, and reduces downtime. Integrating AI and ML techniques into DTs offers a comprehensive, dynamic, and accurate representation of the AM process and significantly improves efficiency and reliability. \textbf{Table~\ref{tab:dt_metal_am}} outlines key parameters, AI/ML techniques, tools/software, end goals, and their role in achieving DTs in metal AM.

\begin{table}[h]
\centering
\caption{DTs in metal AM (powder bed usion)}
\label{tab:dt_metal_am}
\begin{tabular}{p{2.5cm}p{2cm}p{3cm}p{3cm}p{3cm}}
\toprule
\rowcolor{gray!20} \textbf{Parameters} & \textbf{AI/ML Techniques} & \textbf{Tools/Software} & \textbf{End Goals} & \textbf{How It Achieves DT} \\
\midrule
Powder Properties & Regression, Clustering & Python (scikit-learn, TensorFlow), MATLAB & Predict powder behavior, classify powder types & Ensures accurate simulation of material properties \\
\midrule
Machine Settings & Optimization Algorithms, Genetic Algorithms & Python (SciPy, DEAP), MATLAB & Optimal machine parameters for quality control & Simulates various machine settings for optimal performance \\
\midrule
Thermal History & Physics-based Neural Networks, Constrained Optimization & TensorFlow, PyTorch & Accurate thermal profile prediction & Captures thermal characteristics affecting material properties \\
\midrule
Mechanical Properties & Reinforcement Learning, Physics-Informed Neural Networks & TensorFlow, PyTorch, COMSOL & Predict mechanical properties of printed parts & Integrates physics-based constraints to ensure realistic mechanical behavior \\
\midrule
Process Monitoring & Real-Time Data Analysis, Anomaly Detection & Python (Pandas, NumPy), MATLAB, Azure IoT & Detect and correct anomalies during printing & Real-time feedback loop for continuous process improvement \\
\midrule
Layer-by-Layer Analysis & Convolutional Neural Networks & TensorFlow, Keras & Analyze each layer for defects & Ensures quality control at each layer of manufacturing \\
\midrule
Post-Processing Effects & Simulation Models, Finite Element Analysis & ANSYS, Abaqus, Python (SimPy) & Predict impact of post-processing steps & Integrates post-processing steps into the DT for end-to-end accuracy \\
\midrule
Sensor Data Integration & Sensor Fusion, Time-Series Analysis & Python (tslearn), MATLAB, Azure IoT & Integrate data from multiple sensors & Provides a holistic view of the process using multi-sensor data \\
\midrule
Predictive Maintenance & Predictive Analytics, ML & Python (scikit-learn), Azure ML & Schedule maintenance before failures occur & Extends the lifespan of the equipment by predicting failures \\
\midrule
Simulation and Validation & DT Frameworks, Simulation Software & MATLAB, Simulink, ANSYS, Abaqus & Validate simulation against real-world results & Ensures the DT is an accurate replica of the physical system \\
\bottomrule
\end{tabular}
\end{table}
\section{Development and Implementation of DTs}~\label{dt1}
The development and implementation of DTs in AM have significantly advanced the field by efficiently identifying defects, optimizing processes, and improving overall print quality. For instance, in Laser-Directed Energy Deposition (L-DED) processes, a Multisensor Fusion-based Digital Twin (MFDT) integrates data from various sensors, such as coaxial melt pool vision cameras, microphones, and infrared thermal cameras, to predict product quality and detect defects like keyhole pores and cracks~\cite{chen2023multisensor}. This method not only enhances defect prediction but also sets the stage for self-adaptive hybrid processing strategies that integrate machining with AM for defect removal and quality improvement.

\begin{figure}[h]
    \centering
    \includegraphics[width=\textwidth]{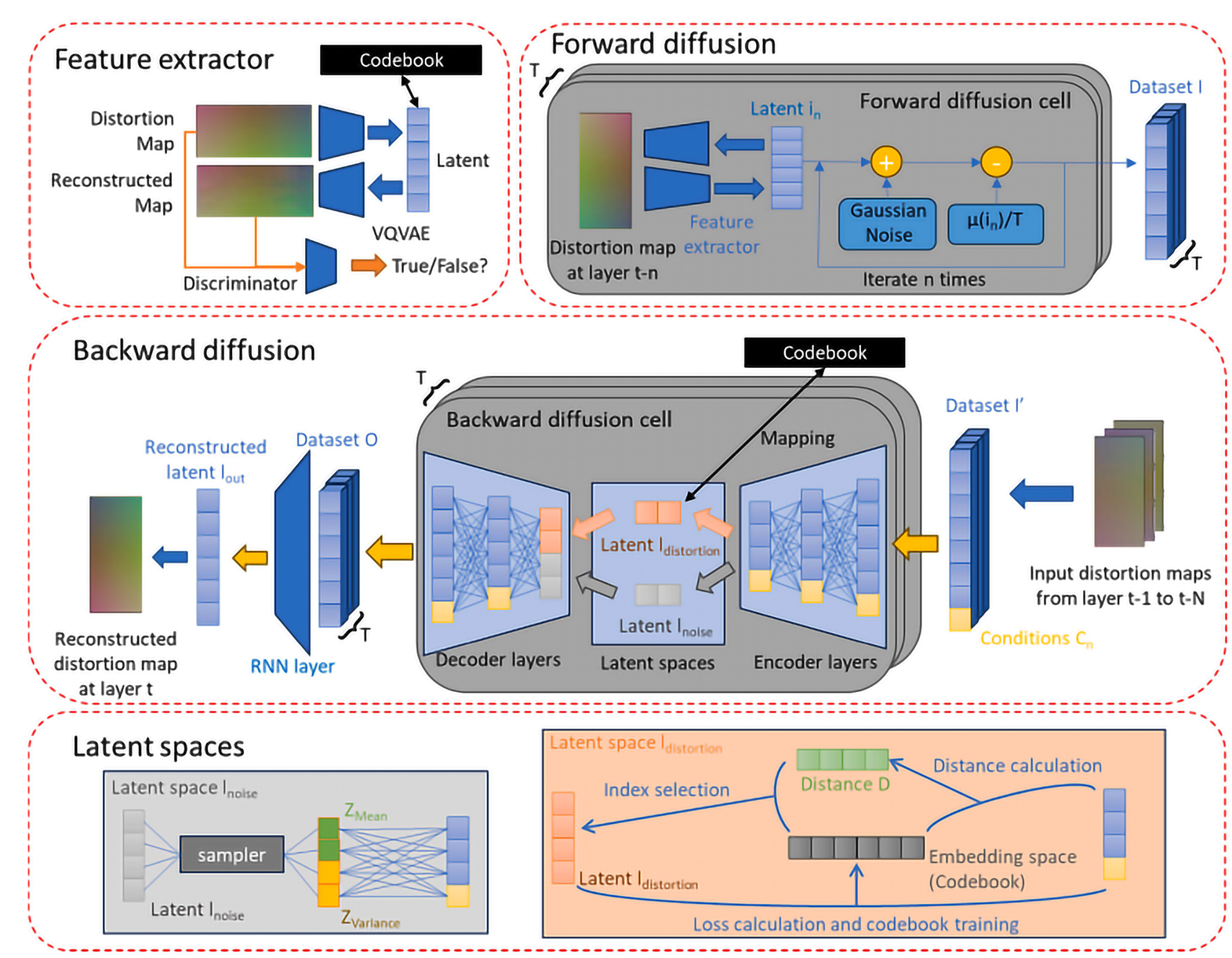}
    \caption{\textbf{The online adaptive simulation model combines a VQVAE-GAN for feature extraction with an RNN for predictions. It predicts future distortion maps from historical data using 2D convolutional layers for spatial features and a recurrent structure for temporal features. Input distortion patterns can come from FEM simulations, model predictions, or sensors. Training minimizes reconstruction errors and maximizes classification accuracy using mean squared error and binary cross-entropy loss functions~\cite{mu2024online}.
}}
    \label{fig:vqv}
\end{figure}

Mu et al. (2021) proposed an adaptive online simulation model to predict distortion fields in Wire Arc Additive Manufacturing (WAAM). This model combined a Vector Quantized Variational AutoEncoder and a Generative Adversarial Network (VQVAE-GAN) for spatial feature extraction with a Recurrent Neural Network (RNN) for time-scale fusion. The system was pretrained using Finite Element Method (FEM) simulated distortion fields (Figure~\ref{fig:vqv}). By using laser-scanned point clouds, the model predicted distortion fields with a Root Mean Square Error (RMSE) below 0.9 mm, outperforming FEM by 143\% and ANN methods by 151\%~\cite{mu2024online}. However, the study only focused on predicting distortion fields in metallic AM. Therefore, this approach might not generalize to other types of AM methods, which might require further investigation. Additionally, the reliance on pretrained FEM-simulated distortion fields for model training might have introduced biases based on the specific conditions and assumptions of the FEM simulations. Furthermore, the experimental validation on thin-wall structures might not fully represent the distortion behavior in more complex geometries or larger parts, potentially limiting the model's applicability in all AM scenarios.

Osho et al. (2022) proposed a 4 Rs framework for creating general-purpose, modular DTs, which consist of four phases: representation, replication, reality, and relational. The representation phase involved understanding and describing the physical system's significant features. Replication duplicated chosen components in a virtual environment. The reality phase employed ML to create an autonomous virtual device for predictions and optimizations. The study demonstrated the representation phase using a FDM machine with temperature and position sensors, validating the framework's precision. This framework aims to simplify DT development from conceptualization to maturity using low-cost sensors and software~\cite{osho2022four}. However, the study only focused on the representation phase and did not cover the other three phases—replication, reality, and relational—in depth. The evaluation of precision in representing the FDM machine was conducted in a controlled environment, which might not fully capture real-world variability and challenges that could affect the DT model's performance under different conditions.

Oettl et al. (2023) combined AM with digital concepts like DTs and Digital Part File (DPF) to enhance AM competitiveness. The study presented a quantitative approach to evaluate the monetary value of using a DPF in the early lifecycle phase of AM production. By analyzing the AM production process, a DPF was translated into monetary value. The study identified 12 aspects describing the benefits of the DPF, allowing a monetary assessment, with an exemplary use case estimating an added value of €20.81. The study highlighted the need for sustainable information in the DPF to generate additional value. However, the study's approach may be limited by the accuracy of cost estimation methods and the applicability of identified benefits across different AM contexts. Additionally, the integration of sustainable information required further development and validation to fully realize its potential benefits~\cite{oettl2023method}.

Yi et al. (2021) applied DTs in AM using AR for a material extrusion printer. They introduced Volume Approximation by Cumulated Cylinders (VACCY) to simulate component geometry. The study integrated electricity use, manufacturing cost, greenhouse gas emission, and energy consumption into the AR-based twin. Tests with three components indicated that AR was suitable for AM DTs and that VACCY effectively simulated printing~\cite{yi2021process}. However, the study demonstrated that the proposed DTs performed better on a desktop AR platform compared to a mobile AR platform. The Frames Per Second (FPS) on mobile phones dropped to 3-6 during the pilot test, which indicated potential performance issues on certain devices. Furthermore, the proposed approaches did not examine the long-term performance stability of the AR-based DTs, which raised questions about their sustainability over extended periods.

Anderson et al. (2021) proposed Time-Driven Activity-Based Costing (TDABC) with DTs to optimize time and equipment capacity in AM. This method analyzed system dynamics and responses, enabling efficient process flow and cost reduction. TDABC calculated costs based on standard time and equipment usage, while DTs modeled time variations to predict costs and identify bottlenecks. This approach helped schedule jobs in advance, optimized machine capacity, and reduced costs~\cite{anderson2021time}. However, the accuracy of stochastic models and the complexity of integrating TDABC with DTs required significant investment and expertise.

\textbf{Table~\ref{tab:devimpl}} summarizes some of the referenced literature that develops and implements various DTs in AM.

\begin{small}
\begin{longtable}{p{.07\linewidth} p{.13\linewidth} p{.16\linewidth} p{.15\linewidth} p{.3\linewidth}}
\caption{Development and implementation of DTs based on the referenced literature. Acc: Accuracy, RMSE: Root Mean Squared Error, AP: Average Precision. Best results are highlighted in bold.}\label{tab:devimpl} \\
\toprule
\rowcolor{gray!20} \textbf{Ref.} & \textbf{Approaches} & \textbf{Applications} & \textbf{Evaluations} & \textbf{Limitations} \\
\midrule
\endfirsthead
\caption[]{Development and implementation of DTs (cont.).} \\
\toprule
\rowcolor{gray!20} \textbf{Ref.} & \textbf{Approaches} & \textbf{Applications} & \textbf{Evaluations} & \textbf{Limitations} \\
\midrule
\endhead
\midrule
\endfoot
\bottomrule
\endlastfoot

\cite{osho2022four} &  4Rs framework for DTs development: Representation, Replication, Reality, and Relational & FDM modeling of AM considering temperature and position sensor & Acc: 95\% & Lack of real-world applications and the scalability. \\ \midrule

\cite{oettl2023method} & Integrates digital concepts with DPF & DPF evaluation & Monetary value: €20.81 & Did not consider modularity and extensibility and had incomplete lifecycle consideration. \\ \midrule

\cite{yi2021process} & AR-based DTs, VACCY & Process monitoring and optimization & Mean porosity: 7.57\%, FPS: 3-6 & The volume of the virtual object may vary from the volume of the real object, and high computational power is needed.\\ \midrule

\cite{mu2024online} & Diffusion Model, VQVAE-GAN, RNN & Online simulation for distortion prediction in WAAM & RMSE: <0.9 m; Outperformed FEM by 143\%, ANN by 151\% & The need for improving the model's intrusiveness, performance on complex geometries, and the effectiveness of dimension reduction techniques has not been thoroughly investigated, and there is high computational complexity. \\ \midrule

\cite{anderson2021time} & TDABC based DTs & Cost optimization in AM processes & - & High initial setup costs. \\ \midrule

\cite{reisch2023prescriptive} & Prescriptive process parameter adaptation & Fault-tolerant manufacturing in WAAM & Defect compensation: 93.4\% & Lack of empirical data to validate the proposed models and methods, did not address the potential complexities and uncertainties involved in creating and maintaining accurate DTs, and needed more precise parameter optimization. \\ \midrule

\cite{xu2024augmented} & AR-Assisted Cloud AM (AR-CAM) & Integration of advanced computational and manufacturing technologies such as DTs, AR, and Cloud Additive Manufacturing (CAM) & Improved cooperation efficiency & Focused on a single case study and did not consider scalability issues. \\ \midrule

\cite{attariani2022digital} & Synchronized circular laser array in PBF & Microstructure control in Ti–6Al–4V alloy & Equiaxed volume fraction: 45\% at 500 W & Did not offer optimal overlap guidelines or consider environmental factors like temperature and humidity.
 \\ \midrule

\cite{sampedro20233d} & Ensemble 3D-AmplifAI algorithm & Fault monitoring in 3D printers & Acc: 82.35\%, Precision: 85.71\%, F1-score: 80\% & Needs more sensors for real-time decisions. \\ \midrule

\cite{scime2022scalable} & Scalable Cyber-Physical Systems & Component performance prediction & Data from 410 builds on seven powder bed printers & High variability in processes, and limited focus on physical testing. \\ \midrule

\cite{mourtzis2021digital} & DTs for FDM & Process monitoring and optimization & Improved reliability and process optimization & Challenges of database and module integration, complexity of optimizing process parameters, and lack of information regarding the human operator and DTs. \\ \midrule

\cite{henson2021digital} & DTs for distortion detection & Failure detection in FDM machines & Rapid failure detection in 2 of 3 test prints & Inability to detect small errors and lack of in situ detection for certain errors. \\ \midrule

\cite{henson2021digital} & DTs for distortion detection & Failure detection in FDM machines & Rapid failure detection in 2 of 3 test prints & Inability to detect small errors and lack of in situ detection for certain errors. \\ \midrule

\cite{stavropoulos2022digital} & Data-driven models in DTs & Performance optimization in Laser Powder Bed Fusion (LPBF) & Improved process time and quality & Potential delays in processing the large volumes of data and did not consider the potential mechanical and control challenges associated with coordinating multiple robots in a complex manufacturing environment. \\ \midrule

\cite{bauer2023multi} & DTs with AI-based error prevention & Quality monitoring in LPBF & Detection accuracy: 91\% & Dependence on simulated data, sensor setup complexity, and environmental factors. \\ \midrule

\cite{sieber2020enhancement} & DTs for optical precision & Enhancing 3D printing for optical surfaces & Improved process control and precision & Insufficient form fidelity in 3D inkjet printing, dependence on measurement data, and the need for integrating surface roughness measurements. \\ \midrule

\cite{jyeniskhan2023integrating} & DTs with ML & Real-time monitoring and defect detection & AP: 92\%; Defective objects: 91\%, Non-defective objects: 94\% & Limited to access through the same wireless network, dependency on image quality, limited middleware architecture, and constant changes in the IP address of the Raspberry Pi, making remote access difficult. \\ \midrule

\cite{li2022time} & Reduced Gaussian Process Emulator (RGPE) and Sketched Emulator with Local Projection (SELP) & Temperature profile prediction, thermal history emulation & Reduced GP emulator: Acc: 95\% for 99.38\% of tests, time: 0.036s; SELP: Acc: 97.78\%, time: 42.23s & RGPE: controlled accuracy; SELP: high accuracy, longer time. \\ \midrule

\cite{castello2024multiscale} & Mean-Field Homogenization and Finite Element Analysis & DTs development with topology optimization & MF Single-layer: Elastic modulus error: -0.201 GPa, FE Multi-layer: Elastic modulus: 9.000 GPa, discrepancy: 7.8\% & Discrepancy in elastic modulus, focus on specific composite material, high computational cost, and early-stage DTs challenges. \\ \midrule

\cite{slepicka2022digital} & AM and Fabrication Information Modeling & Automation of Quality Control in AM processes & - & Dependency on human supervision, unrealized vision for architecture, engineering, and construction industry. \\ \midrule

\cite{barnowski2022multifunctional} & Multifunctional laser technology & Agile production technology integration & High initial setup complexity & High setup complexity, needs advanced simulation. \\ \midrule

\cite{franciosa2022digital} & In-line measurement systems & Assembly process optimization for aircraft panels & MSE: 0.08 mm, Time saving: 75\% & High investment in smart technologies. \\ \midrule

\cite{furuya2022digital} & Digital Triplet with AM & Complex structures in thermal-hydraulic applications & - & Limited practical applications, lacked detailed methodology and did not provide detailed information on how these systems could be scaled up or down for different applications. \\
\end{longtable}
\end{small}

\section{Process Improvement and Hybrid Manufacturing}~\label{dt2}
DTs are revolutionizing process improvement and hybrid manufacturing by enabling precise simulations and real-time adjustments. In process improvement, DTs help identify inefficiencies, predict maintenance needs, and optimize workflows, which leads to significant cost savings and increased productivity. In hybrid manufacturing, DTs integrate various manufacturing techniques, allowing for seamless transitions between processes. They support the fine-tuning of complex operations, ensure optimal resource allocation, and improve the quality of final products. By leveraging DTs, manufacturers can achieve higher flexibility and efficiency in their production systems~\cite{papacharalampopoulos2023integration,hartmann2024digital,montoya20212d}.

Papacharalampopoulos et al. (2023) presented a framework that integrated process selection with DT capabilities and Industry 5.0 (I5.0) criteria, with a focus on sustainability and human inclusion. The study evaluated AM and laser welding and demonstrated that I5.0 criteria improved worker well-being and production efficiency. Additionally, the integration of I5.0 Key Performance Indicators (KPIs) improved productivity, material usage, and energy efficiency across all scenarios~\cite{papacharalampopoulos2023integration}. However, the work focused on a case study evaluating only two parts using AM and laser welding which might not capture the full spectrum of manufacturing processes. The economic implications of implementing the proposed framework were not fully explored. Furthermore, the current framework might not fully account for the nuances of manufacturability and quality in process selection.

Montoya et al. (2021) analyzed the influence of laser intensity distribution, spot radius, and process speed on the thermal history of a substrate in laser-based AM. The study examined gaussian, uniform circular, and uniform rectangular laser power distributions in 2D linear substrate heating simulations, using laser radii of 2.0 mm, 2.5 mm, and 3.0 mm. It was found that these parameters significantly influence the maximum temperature and width and depth of the heat-affected zone. Due to simplicity and acceptable temperature error ~\cite{montoya20212d}, linear simulations were sufficient for DT-based design. However, this work did not consider any heat phenomena in the z direction, which is not appropriate for all processes of laser heating. In the proposed approaches, substrate properties such as specific heat, density, and thermal conductivity were considered constant. However, these properties can vary with temperature and affect the accuracy of the simulation. Moreover, the simulations were linear; nonlinear effects, including those of temperature-dependent material properties, convection, radiation—all these together may play an important role in substrate thermal behavior and thus significantly affect the accuracy of simulations—were not taken into account.

\begin{figure}
    \centering
    \includegraphics[width=\textwidth]{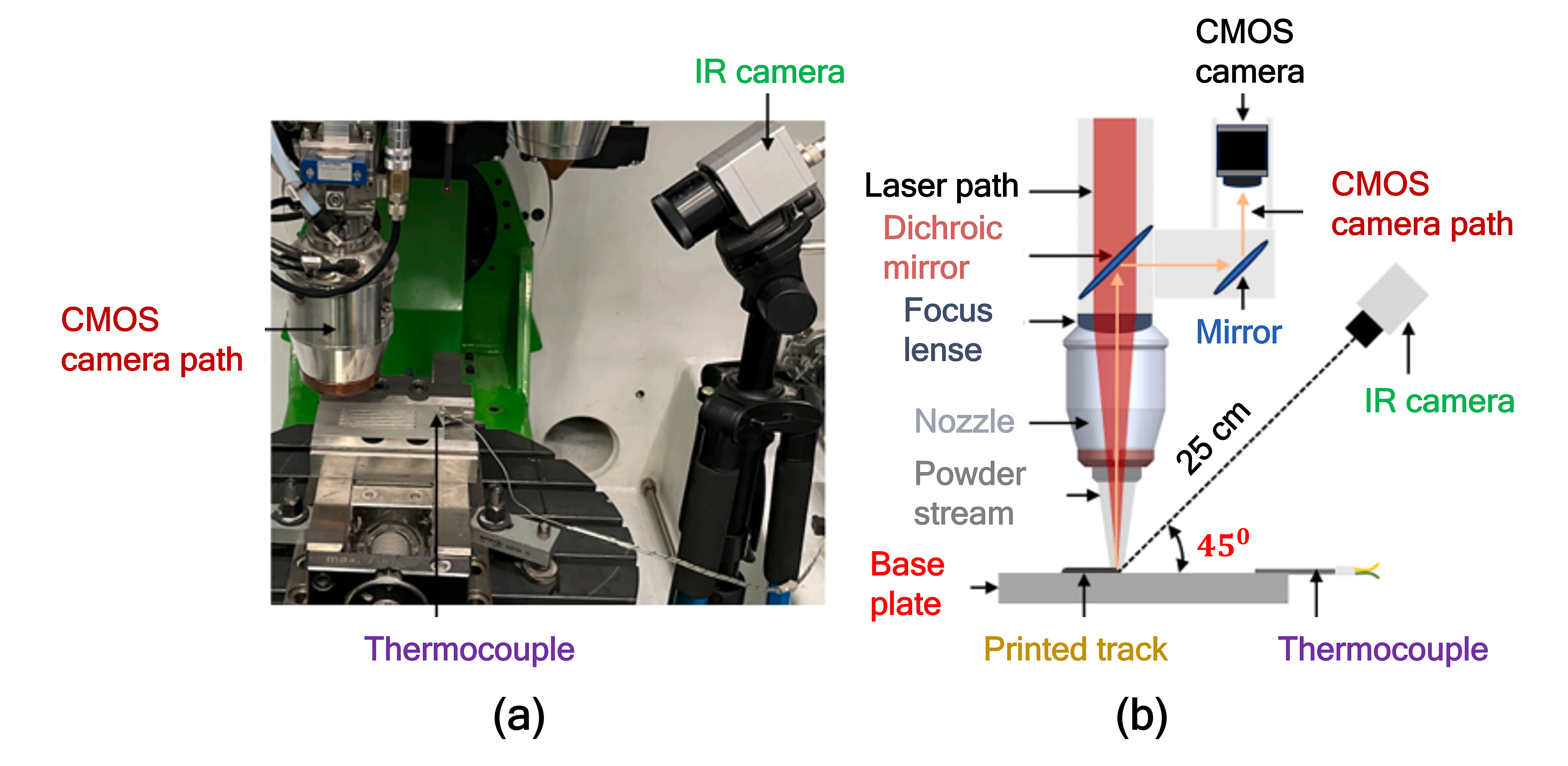}
    \caption{\textbf{Monitoring setup: (a) sensor systems inside the building chamber, and (b) schematic front-view of the monitoring setup in a BeAM Modulo 400 L-DED system. The system features a SINUMERIK ONE control unit, a 2 kW Ytterbium fiber laser (IPG Photonics YLR2000), and a build volume of 400 × 400 × 400 mm³ with two deposition heads. The IR camera is positioned 25 cm from the base plate center at a 45° angle to the laser beam for optimal coverage~\cite{hartmann2024digital}}.}
    \label{fig:thermo}
\end{figure}

Hartmann et al. (2024) studied a multiscale DT for L-DED to reduce time and money in AM. The authors integrated the global model with the local models to simulate laser-powder interactions and high cooling rates successfully on this multiscale DT. A detailed monitoring set-up tracing—and thus validating the elements—was used for this L-DED process, as shown in Figure ~\ref{fig:thermo}. The local model predicted clad dimensions with an error of less than 11.2\% and maximum temperatures with 11.9\% deviation. The global model's mean temperature errors ranged from 2.79\% to 6.81\%. Overall, the DT demonstrated a mean error of 4-5\% for melt pool dimensions and 7\% for temperatures~\cite{hartmann2024digital}. Although the DT was validated through experimental testing, the validation was limited to specific conditions, materials, and geometries, which might not provide the full spectrum of AM in real-world applications. The study also implicitly acknowledged the shortcomings of controlling computational efficiency.

Beckman et al. (2021) presented a methodology for creating and modifying DT replicas of nonwoven fibrous air filtration media using Python scripting and Ansys SpaceClaim. This approach replicated the fiber geometry, enabling computational modeling of air filtration. The study analyzed the effect of fiber stiffness on the filter's solid volume fraction and thickness and modeled contemporary air filtration media to gather data on airflow resistance and particle capture efficiency. The Single Fiber Efficiency (SFE) model estimated performance, and the generated DT fibrous geometry compared well with Scanning Electron Microscopy (SEM) imagery. The study demonstrated the utility of the SFE and pressure drop models in predicting filtration efficiency and flow resistance. A high-efficiency particulate air-quality DT filter slice was analyzed, showing good agreement with SEM images~\cite{beckman2021digital}. However, the proposed approaches needed to refine the DT geometry creation algorithm and conduct Computational Fluid Dynamics analysis to match experimental, analytical, and computational models. Additionally, extending the method to other computational models, such as those for mechanical and thermal evaluation of fiber-reinforced composites, was necessary to improve the accuracy and applicability of DT models in various nonwoven fiber mesh computational modeling applications.

Sofic et al. (2022) investigated the transformation of manufacturing firms through digital services, using data from 136 Serbian firms. The study employed social network analysis, correlation analysis, and interviews. Results showed that firms using technologies like AM, big data analytics, and DTs achieved higher annual turnover, especially during COVID-19. The study underscored the importance of profit-oriented strategies and smart manufacturing technologies for resilient and financially robust processes and identified sectors and technologies that boosted turnover. While confirming findings from developed countries, it added evidence from developing countries on digital transformation's role in improving sustainability and resilience~\cite{sofic2022smart}. However, the study was solely based on a single dataset, which might not be a feasible approach to draw a general conclusion about DTs. Additionally, the study focused only on financial resilience, without considering environmental and engineering resilience as potential factors.

Reisch et al. (2023) proposed a smart manufacturing system for first-time-right printing in Wire Arc Additive Manufacturing (WAAM) by compensating for defects using DTs. The system predicted the future position of the welding torch, analyzed its spatial context, and adjusted process parameters to correct defects, enabling a fault-tolerant manufacturing process. It successfully compensated for discontinuity defects in 93.4\% of cases~\cite{reisch2020robot}. However, the WAAM process struggles to achieve high accuracy. The surface quality produced by the WAAM process does not meet the precision requirements for functional surfaces. This limitation necessitates additional reworking steps, such as milling, to achieve the desired surface characteristics, which can increase production time and costs.

Turazza et al. (2020) investigated the integration of digital technologies in manufacturing, focusing on AM polymer extrusion dies using continuous liquid interface production. The study highlighted potential time and cost reductions compared to conventional methods, with polymer dies successfully producing up to 1000 meters of products. Profile polymer AM dies were created based on existing steel dies and demonstrated robustness and met the required specifications, although they were not yet comparable to the reference sample. Autodesk Moldflow was adapted for flow simulations, which showed good correspondence with real products and aided in the die design phase~\cite{turazza2020towards}. However, the quality of extruded products did not match reference samples, and further evaluation was required to validate a simulation model that accurately predicted melt flow behavior. Addressing these issues was essential for optimizing die design and fully realizing the benefits of integrating AM polymer dies in manufacturing processes.

Table~\ref{tab:dthybrid} summarizes some of the existing literature that emphasizes integrating DTs with hybrid manufacturing concepts.
\begin{small}
\begin{longtable}{p{.07\linewidth} p{.13\linewidth} p{.15\linewidth} p{.15\linewidth} p{.30\linewidth}}
\caption{DT and hybrid manufacturing based on the referenced literature.}\label{tab:dthybrid} \\
\toprule
\rowcolor{gray!20} \textbf{Ref.} & \textbf{Approaches} & \textbf{Applications} & \textbf{Evaluations} & \textbf{Limitations} \\
\midrule
\endfirsthead
\caption[]{DT and hybrid manufacturing (cont.).} \\
\toprule
\rowcolor{gray!20} \textbf{Ref.} & \textbf{Approaches} & \textbf{Applications} & \textbf{Evaluations} & \textbf{Limitations} \\
\midrule
\endhead
\midrule
\endfoot
\bottomrule
\endlastfoot

\cite{papacharalampopoulos2023integration} & DT with I5.0 criteria & Automated process selection and scheduling & Increased productivity, energy efficiency, and profitability with I5.0 support & Limited to a case study of two parts, focuses only on AM and laser welding, and may not fully address manufacturability and quality factors. \\ \midrule

\cite{malim2023structural} & Linear isotropic numerical model & Structural design for high-altitude UAVs & Maximum error: 6\% (tensile), 2.44\% (bending); Safety factors: >2 (take-off), >5 (cruise) & Focused only on tough Polylactic Acid (PLA) and Acrylonitrile Butadiene Styrene (ABS), used a linear isotropic model, may not capture complex material behavior, and did not consider the effects of UV, temperature, and humidity on material properties. \\ \midrule

\cite{montoya20212d} & 2D linear transient thermal model & Laser-based AM & Reasonable temperature error compared to non-linear Finite Element Analysis & Neglected heat transfer in the z-direction, assumed constant substrate properties, used only linear simulations; ignored non-linear effects, and omitted temperature-dependent material changes. \\ \midrule

\cite{ferreira2022physics} & Inherent Strain Method & Predicting distortions in LPBF & Faster simulation times with GPU; deformation predictions in good agreement with reference results & Limited discussion on GPU cost and accessibility, potential accuracy trade-offs with inherent strain method, unclear scalability for larger, more complex parts, and insufficient validation against experimental data. \\ \midrule

\cite{hartmann2024digital} & Multiscale DT & Process optimization in AM & Clad dimensions error: <11.2\%, max temperature deviation: <11.9\% & Model specific to DED-L process, limiting broader applicability and unclear generalizability across diverse materials and geometries.\\ \midrule

\cite{kruckemeier2022concept} & DT-based virtual part inspection & Holistic quality assurance for AM & Conceptual architecture for virtual part inspection & Lacks specific implementation guidance for real-world application, validation limited to a desktop Fused Layer Modeling (FLM) printer, may not represent industrial-scale processes, and DT architecture needs more precise adaptation for virtual part inspection. \\ \midrule

\cite{beckman2021digital} & DT replica creation & Modeling nonwoven nanofibrous filters & Good agreement with SEM imagery; filtration efficiency and flow resistance estimation & Limited exploration of applicability beyond air filtration media and need for validation in diverse nonwoven fiber mesh applications.\\ \midrule

\cite{ertveldt2020miclad} & Closed-loop controlled laser metal deposition & Real-time monitoring for AM & Detailed mapping of deposition geometry and temperature & Findings may not generalize beyond the specific MiCLAD machine, lack discussion on spatial resolution and measurement accuracy, and potential perception issues due to template errors in the original output. \\ \midrule

\cite{reisch2020robot} & Robot-based WAAM setup & Production of large-scale metal parts & Best results with 0° tilt and 0°, -5°, -15° lead angles & Focuses mainly on torch orientation; framework not yet fully developed for anomaly detection and control optimization; limited accuracy and component complexity for functional surfaces. \\ \midrule

\cite{sofic2022smart} & Mixed-method approach & Impact of digital technologies on manufacturing firms & Positive correlation between digital technologies and turnover & Limited to Serbian dataset; focuses only on financial resilience. \\ \midrule

\cite{turazza2020towards} & Continuous liquid interface production & Fabrication of polymer extrusion dies & Successfully produced up to 1000 m of conformal products & Needs evaluation for precise melt flow prediction; quality below conventional samples. \\

\end{longtable}
\end{small}

\section{Integration with Industry 4.0 Technologies}~\label{dt3}

By integrating Industry 4.0 technologies in AM, DTs can play a major role in driving advancements in smart manufacturing. By linking physical assets with their digital counterparts, DTs enable seamless integration of IoT devices, AI, ML, and big data analytics. This connectivity allows real-time data collection and analysis and improves decision-making and operational efficiency. In addition, support for predictive maintenance, advanced simulations, and automated control systems increases productivity and reduces downtime. As a result, integrating DTs with Industry 4.0 technologies leads to smarter, more responsive, and adaptive manufacturing processes~\cite{tao2018digital}.

Assuad et al. (2022) proposed a framework for de- and remanufacturing systems within the circular economy, using Human-Cyber-Physical Systems (HCPS) and AM to refurbish damaged parts. The framework, demonstrated with the CP factory of MANULAB at NTNU, integrated CPS and DT environments to automate operations like disassembly and quality assessment. This approach aimed to reduce manual labor and improve efficiency~\cite{assuad2022proposed}. However, it depended on advanced technology integration, which could be costly and complex, and further research was needed to optimize and scale the framework for various products and manufacturing contexts.

\begin{figure}
    \centering
    \includegraphics[width=\textwidth]{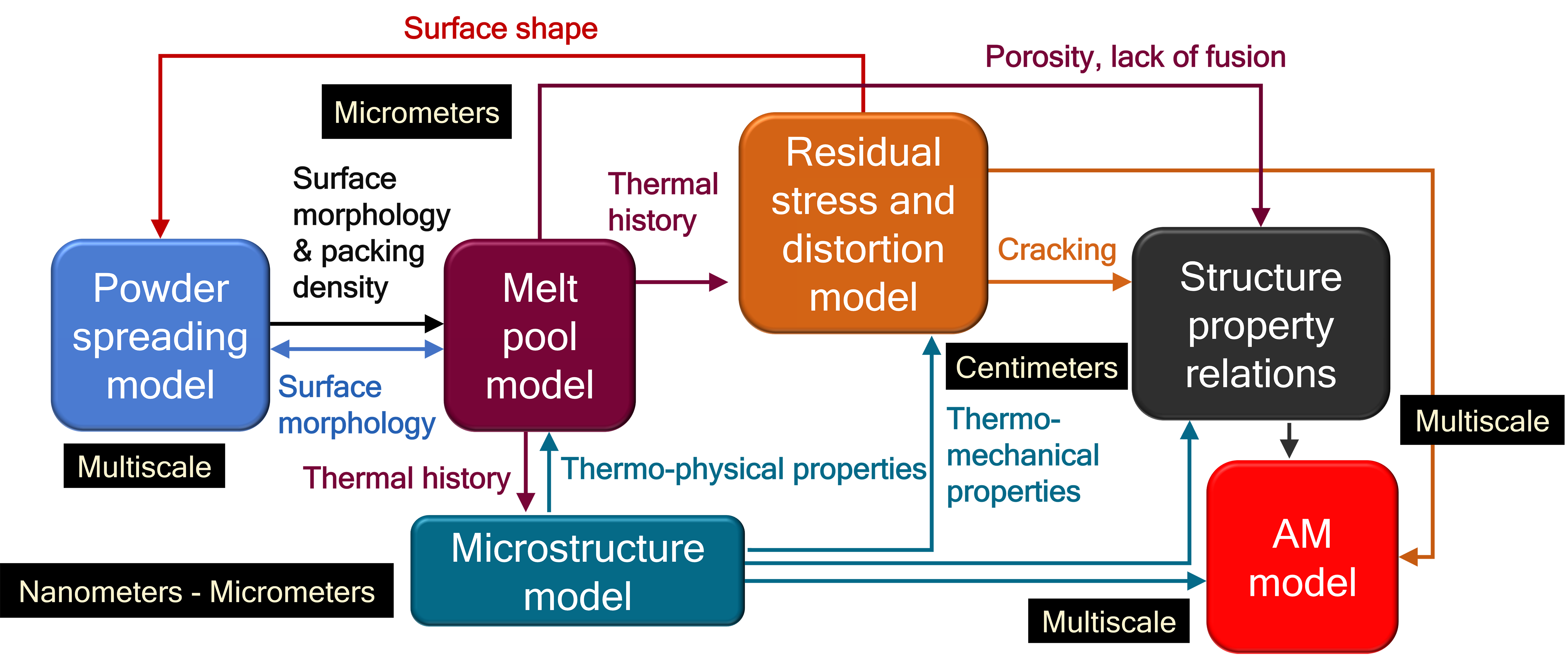}
    \caption{\textbf{The interactions between the individual sub-processes within the powder-bed AM process are examined. The length scale of each sub-process and potential modeling methods are also identified. To maintain clarity, other inputs such as process parameters and thermo-physical data are omitted~\cite{gunasegaram2021towards}.}}
    \label{fig:tmulti}
\end{figure}

Gunasegaram et al. (2021) highlighted the role of AI within DTs to improve productivity and quality in metal AM. Modeling the AM process involved considering various subprocesses that occurred sequentially and/or simultaneously, making it a multiscale problem. The complexity increased as these subprocesses influenced each other, often involving two-way couplings (\textbf{Figure~\ref{fig:tmulti}})~\cite{gunasegaram2021towards}. However, one major limitation of this proposed approach was the difficulty in linking models of different subprocesses across various scales and physics, which was essential for the overall accuracy and functionality of the DTs. Moreover, designing reliable DTs for the AM processes is very challenging due to the scarcity of experimental data. Rapid model development for the solution of real-time problems needs high computing power and extensive specialist knowledge.

Auyeskhan et al. (2023) presented a VR-based decision-support framework for the optimization of assembly design in AM. The research integrated Design for Assembly (DfA) and Design for Additive Manufacturing (DfAM) with Axiomatic Design (AD) theory and exploited VR experiments to prove the independence of Non-Functional Requirements (NFRs), such as assembly time and displacement error. The paper demonstrated the framework on an industrial lifeboat hook. It scored assembly combinations for their probability density to find out those designs that would more likely satisfy NFRs. This framework provided a grounded base to check design independence, quantified NFRs in VR, and made the workflow automatic with the aid of Graphical User Interface (GUI)~\cite{auyeskhan2023virtual}. However, dependence on VR experiments might reduce the actual manufacturability in real-world environments. Furthermore, the few variations in design that were tested may not fully exploit all of the significant potentials of AM, which allows for a much greater range of designs to be prepared.

Lehner et al. (2023) integrated behavioral models into DT platforms, traditionally focused on structural aspects. These models were used for simulation, synthesis, and runtime validation so as to support reasoning about actions and execution history checks. The paper presented modeling patterns that improved DT models with behavioral aspects and runtime data without altering existing DT platform codes. The proposed approaches showed promising results when tested on the Microsoft Azure DT platform in a 3D printing use case~\cite{lehner2023pattern}. One of the main limitations of this study was that behavioral models, which are essential for simulation, synthesis, and runtime validation, were not well integrated into existing DT platforms. This gap hindered the ability to reason about possible next actions or check the execution history effectively.

Dvorak et al. (2022) explored hybrid manufacturing by combining metal AM with Computer Numerical Control (CNC) machining. They addressed problems like part placement, identified the work coordinate system, and reduced the stiffness of preforms, which limited how much material could be removed. They introduced a DT for CNC machining of a WAAM part. The proposed method used structured light scanning to identify the stock model, a fused filament fabrication apparatus to define the coordinate system, frequency response function measurements for stable milling parameters, and post-manufacturing measurements to confirm the design intent. The proposed approaches were validated using audio data and surface profilometry to ensure there was no chatter and the desired surface finish was achieved~\cite{dvorak2022machining}. However, near-net shape designs made part placement and work coordinate system identification more challenging and prone to errors. Additionally, implementing the DT required advanced equipment which might not be readily available in all manufacturing settings.
Table~\ref{tab:indust4.0} summarizes some of the existing literature that emphasizes integrating DTs with Industry 4.0 concepts.
\begin{small}
\begin{longtable}{p{.07\linewidth} p{.15\linewidth} p{.15\linewidth} p{.15\linewidth} p{.3\linewidth}}
\caption{Integration of DT with Industry 4.0 technologies based on the referenced literature. DT: Digital Twin.}\label{tab:indust4.0} \\
\toprule
\rowcolor{gray!20} \textbf{Ref.} & \textbf{Approaches} & \textbf{Applications} & \textbf{Evaluations} & \textbf{Limitations} \\
\midrule
\endfirsthead
\caption[]{Integration of DT with Industry 4.0 technologies (cont.).} \\
\toprule
\rowcolor{gray!20} \textbf{Ref.} & \textbf{Approaches} & \textbf{Applications} & \textbf{Evaluations} & \textbf{Limitations} \\
\midrule
\endhead
\midrule
\endfoot
\bottomrule
\endlastfoot

\cite{assuad2022proposed} & HCPS, DT & Circular economy, refurbishing parts & - & High dependence on manual labor, complex system structure \\ \midrule

\cite{gunasegaram2021towards} & AI, DT, Multiscale-multiphysics models, ML & Improving process productivity and quality & - & Technical hurdles in models; scarcity of experimental data; need for standardization; collaboration challenges \\ \midrule

\cite{auyeskhan2023virtual} & VR, DfA, DfAM, and AD theory & Decision-support for assembly design & - & Does not consider various build orientations; extensive resources needed for Design Rules (DRs) \\ \midrule
\cite{lehner2023pattern} & Behavioral modeling patterns & Augmenting DT models with behavioral models & Enhanced modeling and reasoning & Limited generalizability due to VR-based experiments; restricted design variants tested (only three variations); may not fully capture the complexity of human factors in assembly \\ \midrule

\cite{dvorak2022machining} & DT for CNC machining & Machining of WAAM preforms & Accurate modeling of machining process & Near-net shape designs introduce stringent part placement requirements; reduced preform stiffness may slow down machining process; implementation requires advanced equipment and technical expertise \\

\end{longtable}
\end{small}

\section{Discussions}~\label{dt4}
DTs are a new technology in AM with many uses and benefits. Traditional simulations are often slow and not scalable. DTs use advanced techniques like voxelized geometry and parallel adaptive octree meshes for real-time physics-based simulations. This makes the simulations faster and more efficient. DTs also help detect and correct defects early, using data from various sensors, like acoustic sensors and thermal cameras, for on-the-spot defect detection and process adjustment. This leads to higher efficiency and less waste~\cite{ gamdha2023geometric, chen2023multisensor}.

The integration of IoT sensors boosts the ability of DTs. IoT sensors monitor and diagnose issues in real time, ensure the smooth operation of AM machines, and maintain quality. The metaverse provides a virtual space for the test and simulation of 3D printers. This virtual test is risk-free and cost-effective compared to physical hardware tests. Combined with Deep Learning algorithms, it offers accurate diagnostics and predictions, which improve reliability and efficiency~\cite{duman2023modeling, sampedro2023metaprinter}.

\subsection{Current Techniques Used and Feasibility}
Several techniques are employed to develop and implement DTs effectively:
\begin{itemize}
    \item[\ding{113}] \textbf{Machine Learning and AI.} Machine Learning (ML) and AI are being used in developing DTs for AM. For example, in AM, DTs use binary classification models in environments like Matlab Simulink to detect abnormal conditions that could disrupt operations and affect product quality~\cite{duman2023modeling}. High-precision technologies such as LPBF benefit from DTs that use multi-sensor setups and supervised learning to achieve over 91\% accuracy in detecting printing issues. This reduces the need for physical prototypes and speeds up development times~\cite{bauer2023multi}. Additionally, DTs generate extensive datasets for ML models, significantly cutting the time needed to develop accurate predictive models for production control applications, such as order promising in job shops~\cite{jain2023digital}.
    \item[\ding{113}] \textbf{Augmented Reality (AR).} In AM, AR visualizes the DT of a manufacturing system, allowing operators to interact with the virtual model in real-time and make informed decisions based on data from the DT~\cite{duman2023modeling}. For example, AR supports visualizing the DT on a mobile device, enabling remote monitoring and control of industrial processes. This was demonstrated in the MPS 500 modular production station, where AR displayed real-time data from the DT on a mobile phone. Additionally, AR facilitates live updates of DTs in reconfigurable factory settings, allowing operators to match and update the DTs with their physical counterparts directly on the workstations using an AR authoring tool. This ensures the DTs remain accurate and up-to-date during production changes~\cite{caiza2022digital, sampedro2023metaprinter}.
    \item[\ding{113}] \textbf{Simulation-Based Programs.} These programs are essential for evaluating economic benefits and optimizing production processes within DTs. Simulation-based approaches have been used to assess the financial advantages of Digital Part Files (DPFs) in AM, highlighting how simulations can drive cost-effective decisions and improve the economic viability of AM operations~\cite{oettl2023method, papacharalampopoulos2023integration, montoya20212d}.
    \item[\ding{113}] \textbf{Adaptive Online Simulation Models.} Adaptive simulation models integrated with Neural Networks are vital for predicting complex outcomes like distortion fields in WAAM. These models enable real-time data processing within DTs, ensuring high production accuracy and efficiency. By utilizing advanced algorithms, these models can adapt to changing conditions and provide precise predictions to maintain optimal manufacturing performance~\cite{mu2024online, reisch2020robot, hartmann2024digital}.
    \item[\ding{113}] \textbf{Physics-Informed Machine Learning.} Incorporating physical laws and constraints into ML models within DTs improves the reliability and accuracy of simulations. This approach allows DTs to use both data-driven insights and fundamental physics to predict manufacturing outcomes more effectively, leading to better process control and optimization in AM~\cite{hartmann2024digital, beckman2021digital}.
    \item[\ding{113}] \textbf{Data-Driven Approaches.} By leveraging the extensive data generated during manufacturing, data-driven approaches within DTs identify patterns, optimize workflows, and predict maintenance needs. These methods use big data analytics to continuously improve the performance and reliability of DTs in AM~\cite{sofic2022smart, turazza2020towards}.
    \item[\ding{113}] \textbf{Software and Tools.} Various software and tools support the development and implementation of DTs in AM. For instance, Unity often helps create interactive and immersive DT environments. Tools like Ansys SpaceClaim and Autodesk Moldflow are used for precise modeling and simulation of manufacturing processes. These tools improve the capability of DTs to replicate and optimize real-world manufacturing scenarios~\cite{assuad2022proposed, beckman2021digital, turazza2020towards}.
\end{itemize}

\subsection{Challenges}
Despite their advantages, DTs face several challenges:

\begin{itemize}
    \item[\ding{113}] \textbf{High-Quality Input Data Requirements.} DTs rely heavily on high-quality input data to accurately replicate physical systems and processes. Poor data quality can lead to inaccurate models and predictions, and might compromise the twin's effectiveness in real-time applications~\cite{osho2022four}.
    \item[\ding{113}] \textbf{Complexity in Integration.} Integrating DTs with existing manufacturing systems and processes can be complex and resource-intensive. This integration often requires significant investment in advanced technologies, software, and expertise to ensure seamless operation~\cite{anderson2021time}.
    \item[\ding{113}] \textbf{Scalability Issues.} Scalability remains a major concern, particularly when using low-cost sensors and other budget-friendly technologies. Ensuring that these solutions can scale to accommodate larger and more complex systems without performance degradation is a critical challenge~\cite{osho2022four}.
    \item[\ding{113}] \textbf{Computational Demands.} Techniques like AR and adaptive online simulation models require significant computational power, which can hinder their deployment in resource-constrained environments. For example, implementing AR-based DTs on mobile platforms can drastically reduce frame rates due to computational load~\cite{yi2021process}.
    \item[\ding{113}] \textbf{Validation and Verification.} The accuracy of DT models must be constantly validated and verified against real-world scenarios. This ongoing need for validation poses a challenge, particularly in complex systems where discrepancies between the model and reality can arise~\cite{lehner2023pattern}.
    \item[\ding{113}] \textbf{Technical and Non-Technical Barriers.} The creation of multiscale-multiphysics models and linking subprocesses across scales are technically challenging. Non-technical barriers such as standardization, collaboration difficulties, and the need for international cooperation further complicate the implementation of DTs~\cite{gunasegaram2021towards}.
    \item[\ding{113}] \textbf{Real-Time Data Processing.} Maintaining production accuracy through real-time data processing requires robust and efficient algorithms. Adaptive online simulation models must balance accuracy and computational speed, which is a significant challenge in high-speed manufacturing environments~\cite{mu2024online}.
    \item[\ding{113}] \textbf{Sustainability and Lifecycle Integration.} Integrating sustainability into DTs involves promoting product data exchange throughout the lifecycle and ensuring accurate cost estimation methods. The application of sustainable information in DPFs requires further development to realize its full potential~\cite{oettl2023method}.
    \item[\ding{113}] \textbf{Advanced Technology Dependence.} Frameworks that utilize advanced technologies such as HCPS and AM for refurbishing parts depend on sophisticated, often costly technology integrations~\cite{assuad2022proposed}.
    \item[\ding{113}] \textbf{Physics-Based Model Limitations.} Physics-based models are essential for improving the accuracy of DTs but are often complex and resource-intensive to develop. The dependency on such models can limit the practical application of DTs in various contexts~\cite{osho2022four}.
\end{itemize}
\subsection{Future Scope}
The future scope of DTs in AM provides several research opportunities:
\begin{itemize}
    \item[\ding{113}]  \textbf{Advanced Data Integration.} Future DTs in AM might benefit from more sophisticated data integration techniques by combining IoT sensor data, historical data, and real-time monitoring systems which could increase the accuracy and reliability of DTS.
    \item [\ding{113}] \textbf{Enhanced Computational Methods.} DTs in AM might use advanced algorithms and high computational power to perform complex simulations in real-time. Techniques such as PINNs and ML models could achieve faster and more precise predictions with high computational resources.

\item [\ding{113}] \textbf{Sustainable Manufacturing.} Future research could focus on integrating lifecycle assessment data by optimizing resource usage, and by reducing environmental impact. Sustainable DTs might promote circular economy practices, such as recycling and reusing materials, which are crucial for the AM industry.

\item [\ding{113}] \textbf{Broader Applications of AR and VR.} Future development shall target decreasing computational loads and making the AR/VR interfaces of DTs more user-friendly, intuitive, and immersive to monitor and optimize the processes involved in AM.

\item [\ding{113}] \textbf{Integration with I5.0.} One important development aspect could be the integration of DTs with I5.0. DTs shall be created to ensure worker well-being, increase safety, and raise the level of human-machine interaction in AM environments—probably a necessity for the next generation of smart manufacturing.

\item [\ding{113}] \textbf{Real-Time Adaptive Systems.} DTs in AM could be much more adaptive by enabling real-time adaptation to live data inputs. Future advances in AI and ML algorithms would definitely be required for ensuring fast and correct decision-making procedures on improving the efficiency and flexibility of AM systems.

\item [\ding{113}] \textbf{Improved Cost Models.} More accurate and full cost models of DTs in AM would provide better forecasting and budgeting. This considers every element of production, maintenance, and lifecycle cost component so that more informed economic decisions can be made in the context of AM processes.

\item [\ding{113}] \textbf{Expanded Use Cases.} DT adoption can be extended beyond traditional AM processes to new use cases that involve new materials, complex geometries, and multi-material parts. DT technology should be tailored to specific needs and challenges that exist in these diverse applications in order to extend its reach and efficiency for adoption into AM.

\item [\ding{113}] \textbf{Multi-Scale and Multi-Physics Modeling.} Increasing the ability of DTs in AM to handle multi-scale and multi-physics modeling could improve their accuracy and applicability. Integrating models that can simulate physical phenomena at different scales and across various physical domains might be essential for capturing the complex behavior of AM processes.
\end{itemize}
\section{Conclusion}\label{con}
DTs in AM offer a great opportunity to develop optimal real-time monitoring systems for defect detection and process optimization. DTs provide a detailed understanding of melt pool behavior and defect formation by using multi-sensor fusion, such as in the robotic L-DED process, which traditional single-sensor approaches often miss~\cite{chen2023multisensor}.

Physics-based simulations in extrusion-based AM processes create voxelized geometry representations and transient thermal simulations. These simulations predict heat distribution and optimize print parameters in real time~\cite{gamdha2023geometric}. This method improves print quality, reduces material waste, and lowers computational load, suiting complex geometries. The ability to correct defects on-the-fly by generating auto-tuned process parameters and robot toolpaths increases the efficiency and sustainability of AM processes~\cite{karanjkar2021python, chen2023multisensor}.

Despite these significant advancements, DTs in AM face challenges such as the need for high-quality input data, the complexity of integrating various technologies, and the high computational power required for real-time applications. Scalability issues and the need for physics-based models to improve accuracy are important obstacles. However, potential improvements include better data integration, advanced computational methods, and sustainable manufacturing practices.

Incorporating AR/VR, along with I5.0 principles, could further advance the field. Real-time adaptive systems, improved cost models, expanded use cases, and multi-scale and multi-physics modeling could significantly improve the effectiveness of DTs in AM. Increased collaboration and standardization efforts are essential to ensure the widespread adoption and success of DTs in the industry.

In summary, DTs have the potential to revolutionize AM by improving efficiency, sustainability, and innovation. Addressing the current challenges and leveraging emerging technologies will be key to fully realizing their benefits.

\section*{Acknowledgements}
No funding was received for this work.
\section*{Conflict of interest}
The authors declares no conflict of interest.
\bibliographystyle{unsrt}  
\bibliography{main}

\end{document}